\pdfoutput=1

\documentclass[11pt]{article}

\usepackage{emnlp2021}

\usepackage{times}
\usepackage{latexsym}

\usepackage[T1]{fontenc}

\usepackage[utf8]{inputenc}

\usepackage{microtype}
\usepackage{enumitem}
\usepackage{graphicx}
\usepackage{amsmath}
\usepackage{soul}
\usepackage{xcolor}
\usepackage{etoolbox}

\newbool{in_development}
\setbool{in_development}{false}
\newcommand{\red}[1]{{\color{red}{#1}}}
\newcommand{\redcomment}[1]{\ifbool{in_development}{\red{#1}}{}}

\newcommand{\blue}[1]{{\color{blue}{#1}}}
\newcommand{\bluecomment}[1]{\ifbool{in_development}{\blue{#1}}{#1}}

\title{Text Counterfactuals via Latent Optimization and Shapley-Guided Search}

\author{Quintin Pope \and Xiaoli Z. Fern \\
        School of Electrical Engineering and Computer Science\\ Oregon State University \\ \texttt{popeq,xfern@oregonstate.edu} \\} 

\begin{document}
\maketitle
\begin{abstract}

We study the problem of generating counterfactual text for a classifier as a means for understanding and debugging classification. Given a textual input and a classification model, we aim to minimally alter the text to change the model's prediction. White-box approaches have been successfully applied to similar problems in vision where one can directly optimize the continuous input. Optimization-based approaches become difficult in the language domain due to the discrete nature of text. We bypass this issue by directly optimizing in the latent space and leveraging a language model to generate candidate modifications from optimized latent representations. We additionally use Shapley values to estimate the combinatoric effect of multiple changes. We then use these estimates to guide a beam search for the final counterfactual text. We achieve favorable performance compared to recent white-box and black-box baselines using human and automatic evaluations. Ablation studies show that both latent optimization and the use of Shapley values improve success rate and the quality of the generated counterfactuals.

\end{abstract}

\section{Introduction}
Deep neural networks have achieved state-of-the-art performances for many natural language processing (NLP) tasks~\cite{9075398,ruder-etal-2019-transfer}. When applying such models in real world applications, understanding their behavior can be challenging --- the ever increasing complexity of such models makes it difficult to understand and debug their predictions.
A human can explain why an example belongs to a specific concept class by constructing a counterfactual of an example that is minimally altered but belongs to a different class. Contrasting the original example with its counterfactual highlights the critical aspects signifying the concept class. We study a similar approach to understand deep NLP models' classification criteria. 

Given a classifier and an input text, our goal is to \bluecomment{generate a counterfactual by making} a set of minimal modifications to the text that change the label assigned by the classifier. Additionally, our goal is to understand the model's behavior when processing \textit{naturally occurring} inputs, hence we wish to generate grammatically correct and semantically plausible counterfactuals.

Automatic generation of text counterfactuals has been studied in different settings. \citet{qin2019counterfactual} considered counterfactual story rewriting which aims to minimally rewrite an original story to be compatible with a counterfactual event. Wu et al.~(\citeyear{wu2021polyjuice}) used a fine-tuned GPT-2 model to generate general purpose counterfactuals that are not tied to a particular classification model. \citet{yang2020generating} aim to generate plausible-sounding counterfactuals that flip a classification model's decision for financial texts. 



\bluecomment{Related, textual adversaries also aim to change the model prediction (with modifications resembling natural text). The difference is that adversaries further aim to escape human detection (not changing a human’s classification), whereas counterfactuals do not have such requirement. }



Another line of related work is style transfer \cite{sudhakar-etal-2019-transforming, wang2019controllable, hu2017toward}, which aim to modify a given text according to a target style. It differs from adversary or counterfactual generation in that it seeks to fully change all style-related phrases, as opposed to minimally perturbing a text to change a classifier's decision.

White-box approaches have been widely used to generate adversaries or counterfactuals for vision tasks where the continuous inputs can be optimized to alter model predictions \cite{goodfellow2014explaining,carlini2017towards,Neal_2018_ECCV}. Such optimization based approaches are difficult to apply to language due to the discrete nature of text. We circumvent this difficulty by directly optimizing in the latent space of the input towards the desired classification. We then exploit the language generation capability of pre-trained language models, available for most state-of-the-art NLP models such as BERT \cite{devlin2019bert} or RoBERTa \cite{liu2019roberta}, to generate semantically plausible substitutions from the optimized latent representations.
We further introduce Shapley values to estimate the combinatoric effect of multiple simultaneous changes, which are then used to guide a beam search to generate the final counterfactual. 


Leveraging pre-trained language models to generate alternative texts has been a popular black-box approach in the recent literature on text adversaries \cite{li2020bertattack,garg2020bae,li2020contextualized}. Our work presents a first attempt to combine the strength of white-box optimization and the power of pre-trained language models. While Shapley values have been widely studied for the problem of feature importance \cite{lundberg2017unified,sundararajan2020shapley} and data valuation \cite{jia2020efficient}, this is the first effort demonstrating their usefulness for text generation.

We compare our method to several white-box and black-box baselines on two different text classification tasks. Automatic and human evaluation results show that our method significantly improves the success rate of counterfactual generation, while reducing the fraction of input tokens modified and enhancing the semantic plausibility of generated counterfactuals. We also show through ablation studies that both counterfactual optimization of the latent representations and Shapley value estimates contribute to our method's strong performance.

\section{Proposed Method}

\paragraph{Problem statement.}We are given a text classification model, $M$, an initial input token sequence, $X = \{{{x}}_0, ..., {{x}}_{n-1}\}$, with vocabulary $V$. Model $M$ outputs a classification score $ \hat{y}=M(X)\in (0, 1)$, representing $P(y=1|X)$. Based on the score, a class label $y \in \{0, 1\}$ is assigned. We seek to generate a \textit{counterfactual} of $X$, which is defined as a set of tokens, $X' = \{{{x'}}_0, ..., {{x'}}_{n-1}\}$,  that \bluecomment{differs from $X$ in no more than $C_{max}$ percent of locations, is} grammatically plausible, and leads to a different classification, $y'$. \bluecomment{Here $C_{max}$ is an input parameter for maximum changes allowed, and smaller $C_{max}$ imposes stronger restrictions.} 

Note that our setup assumes binary classification, but can be easily extended to multi-class scenario to generate either targeted (with specified $y'$) or un-targeted counterfactuals (with unspecified $y'$).

\paragraph{Method overview.}Our method consists of three steps. First, we generate a set of candidate token substitutions for each position. Second, we evaluate the capacity of these candidate substitutions to change model classification (individually and collectively), Finally, we construct the counterfactual by beam search.

\subsection{Generating candidate substitutions}
We generate candidate substitutions by first performing latent space optimization and then generate substitutions from the trajectory of latent representations using a language model. 

Given an input token sequence $X = \{x_0, ..., x_{n-1}\}$, we assume model $M$ contains an embedding layer that maps this discrete input sequence into continuous embedding $E = \{e_0, ..., e_{n-1}\}$. The goal is to optimize a sparsely altered ${E^{'}}= \{{e^{'}}_0, ..., {e^{'}}_{n-1}\}$ such that the model will output $y'$, a target class different from $M$'s initial prediction $y$. With slight abuse of notation, let $M({E^{'}})$ denote $M$'s classification score when replacing $E$ with $E'$ as the input embedding, we optimize the following objective.

\begin{equation}
    \displaystyle    \min_{E'} CE(M({}E^{'}), {y'}) + \sum_{j=0}^{n-1}{|{e^{'}}_j - e_j|} 
    \label{optimizeE}
\end{equation}
which minimizes the cross-entropy between $M({E^{'}})$ and the desired $y'$, with a LASSO regularization to favor sparse divergence from the original $E$. 

To reduce the sensitivity to the stopping point of optimization and produce diverse candidates, we optimize $E'$ for $K$ steps and consider the full optimization trajectory $\{E'_k:k=1\cdots K\}$ to generate the candidate substitutions using the pre-trained language model associated with model $M$. 

Directly using the pre-trained language model is problematic because it does not operate in the same latent space as model $M$, whose encoder has been fine tuned for the specific classification task at hand. A simple fix to this problem is to use the fine-tuned encoder of $M$ (which is used to optimize $E'$) and retrain the associated language modeling head.\footnote{Specifically, we retrain the language modeling head using the text for which we intend to generate counterfactuals. In our experiments this only involves 1000 data points, leading to a very fast re-training process.} This produces a language model that operates in the same space as the optimized embedding. 

Specifically, we feed each ${E^{'}}_k$ ($k=1,\dots,K$) through the encoder of $M$ and the retrained language modeling head to generate a logit matrix $T_k$ of size $|V|\times n$, where $T_k(s,t)$ quantifies the likelihood of observing the $s$-th token of the vocabulary at the $t$-th location given the overall context of ${E^{'}}_k$. 

To generate $K$ candidate substitutions for each position $t$, we iteratively process $T_1,\cdots, T_K$, selecting the token with the highest logit score excluding the original $x_t$ and previous selections. Let $\mathcal{S}_t^k$ be the set of candidate substitutions for position $t$ generated at iteration $k$ considering $T_k$, it is computed as follows. 
\begin{equation}
    \mathcal{S}^{k}_t = \mathcal{S}^{k-1}_t \cup  \underset{s\notin \mathcal{S}^{k-1}_t \cup x_t}{\text{argmax}} T_k(s, t)
\end{equation}
At the end of this step, we produce a set of $K$ candidate substitutions for each input position. 


\subsection{Evaluating Candidate Substitutions}
In the second step, we compute a metric that measures each candidate substitution's capacity to change the classification when applied in combination with other substitutions. Toward this goal, we consider Shapley value, which was originally proposed in cooperative game theory \cite{shapley1951value} \bluecomment{and has been used to measure feature importance for model interpretability \cite{lundberg2017unified}}. 

For a multi-player coalition-based game, the Shapley value of a player represents how valuable the player is as a potential coalition member. 
In our context, a coalition $L$ is a set of simultaneous substitutions and value $V(L)$ is measured by $L$'s capacity to change model $M$'s prediction. Let $X_L$ denote the input generated by applying all substitutions\footnote{By definition, $L$ must not contain multiple substitutions to the same location, which will create a conflict.} in $L$ to $X$, and $M(X_L)$ be $M$'s prediction score.
We define $V(L)$ to be $M(X_L)-M(X)$ if we wish to flip a negative prediction and $M(X)-M(X_L)$ otherwise.


The Shapley value of a single substitution $s$ measures the expected marginal value of adding $s$ to a coalition not already containing $s$. To ensure computational tractability, we constrain the size of the coalition to be a fixed value $c_s$. As such, coalitions of any other sizes will have value zero. Conceptually this measures the potential value of substitution $s$ when we modify exactly $c_s$ tokens.  

Under this setting, it is straightforward to show that the Shapley value of a single substitution $s$ can be estimated by the following equation:
\begin{equation}
    SV(s) = \frac{1}{|\mathcal{L}_s|}\sum_{L \in \mathcal{L}_s}^{}{V(L)} - \frac{1}{|\mathcal{L}_{/s}|}\sum_{L \in \mathcal{L}_{/s}}V(L) 
    \label{sv}
\end{equation}
$\mathcal{L}_s$ ($\mathcal{L}_{/s}$) denotes the set of coalitions containing (not containing) $s$ that satisfy the size constraint. 

Fully enumerating $\mathcal{L}_s$ and $\mathcal{L}_{/s}$ to compute Equation \ref{sv} is infeasible in most situations. We use two strategies to improve efficiency. First, we apply filtering to remove unimportant locations from further consideration. We adapt the NormGrad saliency method described by ~\citet{rebuffi2020again} to text and use the following gradient-based saliency score.

\begin{equation}
    \text{saliency}(i) = |(\nabla_{e_i} \hat{y}) \odot e_i|^2
    \label{importancescore}
\end{equation}
where $\nabla_{e_i} \hat{y}$ denotes the gradient of the original classification score $\hat{y}$ with respect to $e_i$, the embedding of the $i$-th token, and $\odot$ represents the Hadamard product (elementwise multiplication).

Our second strategy is to approximate the Shapley values by sampling in the space of allowed substitutions. Suppose we want to evaluate each substitution $w$ times on average and there are a total of $N_s$ substitutions to be evaluated. It is interesting to note that we do not need $N_s\cdot w$ evaluations since each evaluation simultaneously contributes to the estimates of all $c_s$ substitutions that it contains. 

\bluecomment{We apply filtering to consider only the top $C_{max} \times n$ locations, and fix the coalition size to be $50\%$ of that ($c_s=0.5 \times C_{max} \times n$). Each important location contributes $K$ candidate substitutions. For input of length $n$, there are $C_{max} \times K \times n$ total substitutions to evaluate. Because each coalition evaluation covers $0.5 \times C_{max} \times n$ substitutions, to evaluate each substitution $w$ times on average, we need to evaluate $2 \times w \times K$ coalitions, which is independent of $n$ and $C_{max}$.}

\subsection{Constructing the Counterfactual}
In the final step, we search for the optimal subset of substitutions via breadth-first beam search. The search space covers all possible subsets of non-conflicting substitutions, each subset corresponding to a unique candidate counterfactual.

We initialize the beam with the root of the search tree, which is the empty subset. At each iteration, we expand a node in the beam with a successor function returning $b$ successors, each adding a single substitution. For a given search node, denoted by its subset $L$, we construct $b$ successors by selecting $b$ substitutions with the best Shapley values that do not conflict in location with any $s\in L$ or introduce a redundant subset. 

We then evaluate each successor node by applying its substitutions to the original input $X$ and computing model $M$'s output on the resulting $X'$. We rank all successors based on the model's score for the desired class $y'$ minus the fraction of tokens modified by the successor in question and populate the new beam with the top $b$ candidates.

We limit the search depth to be $C_{max} \times n$, constraining our method to never modify more than $C_{max}$ percent of the input tokens. During search, if we generate a candidate that $M$ classifies as $y'$, we stop immediately and return that candidate as our final output.  As such, the time we spend for beam search depends on how quickly we find a successful counterfactual. 

\subsection{Summary of approach}
We summarize our method as text \textbf{C}ounterfactuals via \textbf{L}atent \textbf{O}ptimization and \textbf{S}hapley-guided \textbf{S}earch (CLOSS). CLOSS has three primary hyperparameters: 
$K$, the number of candidate substitutions generated per token locations; $w$, the average number of times we wish to evaluate each substitution;     
$b$, the beam width of the beam search that constructs the final counterfactual. The default values are $K=30$, $w=5$, $b=15$. The impact of these parameters will be explored in the experiments.



\section{Empirical Evaluation}
To evaluate our proposed method, we consider two different text classification tasks: sentiment classification and natural language inferences. 
\subsection{Experimental Setup}

\paragraph{Data sources.} We use the IMDB dataset \cite{maas-EtAl:2011:ACL-HLT2011} for sentiment classification. This is a binary classification dataset based on movie reviews from IMDB. For the natural language inference task, we use the QNLI dataset \cite{rajpurkar2016squad}, which is a binary task derived from the Stanford question answering dataset. Each example contains a question and a context. The classifier must determine if the context answers the question. 

Following the evaluation scheme used by~\citet{li2020contextualized}, we sample 1000 random data points from IMDB of length less than or equal to 100 words as our ``short'' IMDB data. We do not filter the QNLI dataset. The average word counts for short IMDB and QNLI are shown in Table~\ref{tab:modelacc} (row 1).
 

\paragraph{Classification models.}
For each task, we consider two classification models, RoBERTa \cite{liu2019roberta} and BERT \cite{devlin2019bert}, trained by TextAttack \cite{morris2020textattack}. We report the performance of both models in Table \ref{tab:modelacc} (rows 2-3).

\begin{table}[h!]
    \centering
    \begin{tabular}{l| c c}
          &  IMDB short & QNLI \\
         \hline
         Avg. words & 57.4 & 37.2\\ \hline\hline
         RoBERTa acc. &  0.971 & 0.934 \\
         BERT acc. &  0.974 & 0.908 \\

    \end{tabular}
    \caption{Model and dataset statistics}
    \label{tab:modelacc}
\end{table}

\paragraph{Evaluation criteria.}
We consider the following performance metrics that measure the ability of a method to successfully generate counterfactuals and the quality of the generated counterfactuals. 
\begin{itemize}[leftmargin=*]
\addtolength\itemsep{-4mm}
    \item \textbf{Failure rate (\%F)}: the percent of inputs for which the method fails to change the model's prediction.
    \item \textbf{Fraction of tokens changed (\%C)}: the average token modification rate among successfully generated counterfactuals. 
    \item \textbf{BLEU}: the average BLEU score between successfully generated counterfactuals and their original inputs. 
    \item \textbf{Perplexity (P)}: Following \citet{zang-etal-2020-word}, we use the exponentiated language modeling loss of GPT-2 \cite{radford2019language} to compute perplexity to score linguistic plausibility.
\end{itemize}

\paragraph{Baselines.}
\bluecomment{We compare against adversarial baselines because we were unable to find counterfactual methods with open-source implementations. We carefully identified a set of baselines closely related to CLOSS with respect to the methodology, specifically focusing on black box methods that leverage pretrained language models (BERT-Attack, BAE), and a white-box method using gradients and beamsearch (Hot-Flip).} 
Unless otherwise stated, we use the implementations in the TextAttack package \cite{morris2020textattack}. All black-box methods use some saliency measure to prioritize substituting important tokens. While CLOSS estimates saliency from the gradient, the black-box baselines use leave-a-token-out estimates, i.e., by removing or masking a token. \\
\textbf{BERT Adversarial Example} (BAE) \cite{garg2020bae} is a black-box method that generates potential substitutions by masking out input tokens and using the pre-trained BERT language model to suggest replacements.\\ 
\textbf{BERT-Attack} \cite{li2020bertattack} is also a black-box method. It generates substitutions by feeding the entire \textit{unmasked} input into the BERT language model to suggest replacements. \\
\textbf{Textfooler} \cite{jin2020bert} is a black-box method that uses 
    word embeddings by \citet{synembeddings2016counterfitting} to generate substitutions by selecting the vocabulary tokens whose embeddings have the highest cosine similarity with the original token.\\ 
\textbf{Probability Weighted Word Saliency} (PWWS) \cite{ren-etal-2019-generating} is a black-box method that uses WordNet synonyms as potential substitutions to preserve the semantic content.\\
\textbf{HotFlip} \cite{ebrahimi2018hotflip} is a white-box method that uses model gradients to estimate the impact of every possible token substitution on the model's classification and then applies beam search to generate counterfactuals. HotFlip implemented in TextAttack only works for word or character-level classifiers, not WordPiece \cite{6289079} classifiers like RoBERTa and BERT. Hence we implemented our own version of HotFlip\footnote{HotFlip can generate insertion/deletion edits in addition to substitutions. Our implementation only considers substitutions to be directly comparable with our method.} to be exactly comparable to our method. 


\paragraph{Adaptation of adversarial baselines for fair comparison.} \bluecomment{ Adversaries differ from counterfactuals in that they additionally seek to retain the text's ``true'' class, relative to human judgement. In this regard, generating adversaries is more difficult than generating counterfactuals. Here we adapt the (adversary-generating) baselines to generate counterfactuals, thereby allowing fair comparison.}

All \bluecomment{original baseline implementations} employ certain heuristic constraints to preserve the original semantic content \bluecomment{(and thus, true class)} of the input. Most methods require a minimum cosine similarity between the Universal Sentence Encoder \cite{cer2018universal} (USE) representations of the modified text with the original input. Additional heuristics include not substituting stop words and requiring substitutions to have the same part-of-speech as the original. These heuristics directly modify the search space for a generation method, and thus can impact both the success and quality of counterfactual generation. 

For CLOSS and our implementation of HotFlip, we do not employ such heuristics. \bluecomment{Additionally, we created an unconstrained} version of the TextAttack \cite{morris2020textattack} implementations  (denoted by suffix `-U') of all other baselines by removing \bluecomment{the adversarial} constraints. \bluecomment{Arguably,  PWWS-U and TextFooler-U remain more constrained than CLOSS because they only use synonyms (WordNet and embedding based, respectively) for substitutions. However, the search spaces of BAE-U, BERT-Attack-U, and HotFlip are fully comparable to CLOSS.} 

TextAttack by default ignores any input misclassified by the model $M$, because the concept of an ``adversarial'' example does not readily extend to misclassified inputs. For counterfactual generation, we do not have this concern. Hence our evaluation seeks to generate a counterfactual that flips the model's classification regardless of its correctness.

\paragraph{Baseline parameters.} \bluecomment{For HotFlip, we  consider two versions: a default version (HotFlip D) that uses parameters suggested by \citet{ebrahimi2018hotflip} and an optimized version (HotFlip O), where the parameters and search procedure are optimized for performance. See Appendix 1 for details of our HotFlip implementations.} For all other baselines, we use the default parameters from TextAttack, which are the recommended parameters from the original papers.

\begingroup
\setlength{\tabcolsep}{4pt} 
\renewcommand{\arraystretch}{1} 

\begin{table*}[h]
\centering
\begin{tabular}{l | c c c c | c c c c |} 
&\multicolumn{8}{c}{\textbf{IMDB short}}\\ \hline
& \multicolumn{4}{c}{RoBERTa} & \multicolumn{4}{|c|}{BERT}\\
Method & \%F & \%C & BLEU & P  & \%F & \%C & BLEU & P \\
\hline
CLOSS & \textbf{4.2} & \textbf{3.1} & \textbf{0.92} & \textbf{72.4} & \textbf{4.1} & \textbf{2.8} & \textbf{0.93} & \textbf{98.9} \\\hline
HotFlip D & 37.0 & 6.5 & 0.86 & 145 & 22.8 & 5.2 & 0.89 & 140 \\
HotFlip O & 7.1 & 5.1 & 0.88 & 122 & 4.5 & 4.18 & 0.90 & 129 \\
\hline 
BAE & 69.4 & 4.6 & 0.86 & 110 & 67.5 & 4.0 & 0.88 & 136 \\
PWWS & 14.6 & 5.9 & 0.83 & 96\hspace{2mm}  & 14.0 & 4.7 & 0.86 & 125 \\
TextFooler & 22.3 & 6.3 & 0.82 & 91.5 & 31.9 & 5.7\hspace{2mm} & 0.83 & 132 \\
\hline
BAE-U & 16.6 & 5.7 & 0.85 & 107 & 25.1 & 4.9\hspace{2mm} & 0.87 & 141 \\
Bert-Attack-U & 6.3 & 4.4 & 0.90 & 78.2 & 22.2 & 4.7 & 0.88 & 120 \\
PWWS-U & 12.1 & 5.9 & 0.83 & 102 & 11.7 & 4.6 & 0.86 & 134 \\
TextFooler-U & 12.7 & 5.7 & 0.85 & 93.9 & 21.0 & 5.2 & 0.86 & 142 \\ \hline
&\multicolumn{8}{c}{\textbf{QNLI}}\\ \hline
& \multicolumn{4}{c}{RoBERTa} & \multicolumn{4}{|c|}{BERT}\\
Method & \%F & \%C & BLEU & P  & \%F & \%C & BLEU & P \\
\hline
CLOSS & 5.1 & \textbf{3.3} & \textbf{0.92} & 92.4 & 3.5 & \textbf{3.3} & \textbf{0.92} & \textbf{143} \\
\hline
HotFlip D & 18.8 & 4.7 & 0.90 & 130 & 19.1 & 4.4 & 0.90 & 174 \\
HotFlip O & \textbf{3.4} & 4.0 & 0.90 & 125 & \textbf{2.1} & 3.8 & 0.91 & 178 \\
\hline

BAE & 34.6 & 3.7 & 0.87 & 94.4 & 33.2 & 4.0 & 0.87 & 175 \\
PWWS & 22.7 & 4.2 & 0.87 & 95.1 & 14.9 & 4.4 & 0.86 & 184 \\
TextFooler & 19.4 & 4.6 & 0.86 & 90.1 & 13.1 & 4.9 & 0.86 & 176 \\
\hline
BAE-U & 6.8 & 4.2 & 0.87 & 104 & 6.0 & 4.1 & 0.88 & 178 \\
Bert-Attack-U & 6.7 & 4.0 & 0.89 & \textbf{87.1} & 4.9 & 3.8 & 0.90 & 156 \\
PWWS-U & 16.0 & 4.3 & 0.87 & 107  & 8.7 & 4.3 & 0.87 & 201 \\
TextFooler-U & 7.4 & 4.4 & 0.87 & 101 & 4.6 & 4.3 & 0.88 & 180 \\
\hline
\end{tabular}
\caption{
Comparison of CLOSS with baselines on the short IMDB and QNLI data. `U' indicates unconstrained version of the baselines. Our implementation of HotFlip uses exactly the same set of constraints as CLOSS. CLOSS values are averaged over three runs.
}
\label{table:finalimdbbaselines}
\end{table*}

\subsection{Results}
We report the results of all methods for short IMDB and QNLI in Table~\ref{table:finalimdbbaselines}. Note that BERT-Attack restrains manipulations of multi-token words in a manner that is computationally intractable on our datasets; thus we do not report performance for the original BERT-Attack. Here we limit all methods to change no more than 15\% of tokens by setting $C_{max}=0.15$.  \bluecomment{The impact of different $C_{max}$ values will be explored later in Figure~\ref{figure:differentpercentchanges}.}

\paragraph{Comparing to white-box baseline.} The default HotFlip implementation substantially underperforms by all measures. Optimizing the parameters and search procedure for HotFlip leads to greatly improved performance. Comparing CLOSS with HotFlip O, we note that the failure rate of our method is slightly worse for QNLI and slightly better for IMDB short. For both datasets, the counterfactuals generated by CLOSS have fewer modifications and higher BLEU scores. The most striking difference is in the perplexity score, where CLOSS wins by a large margin. This is not surprising as HotFlip does not care about the semantic plausibility of the generated sentences whereas our method uses the language model to propose semantically plausible substitutions. Note that GPT-2 and RoBERTa are cased models, while BERT is uncased. This explains BERT's higher perplexity.

\paragraph{Comparing to black-box methods.} We first observe that the heuristic constraints used by these methods have a drastic impact on the performance. Specifically, by removing these constraints, the failure rates of all methods are much reduced. However, the resulting counterfactuals tend to have lower quality, indicated by increased perplexity. Comparing CLOSS to both variants, we see that our method was able to achieve a highly competitive failure rate with few edits. CLOSS also achieves the lowest perplexity in most cases with the exception of RoBERTa model on QNLI, where BERT-Attack-U have slightly lower perplexity. 


\begin{figure*}[ht!]
\centering
\begin{tabular}{ccc}
\includegraphics[scale=0.48]{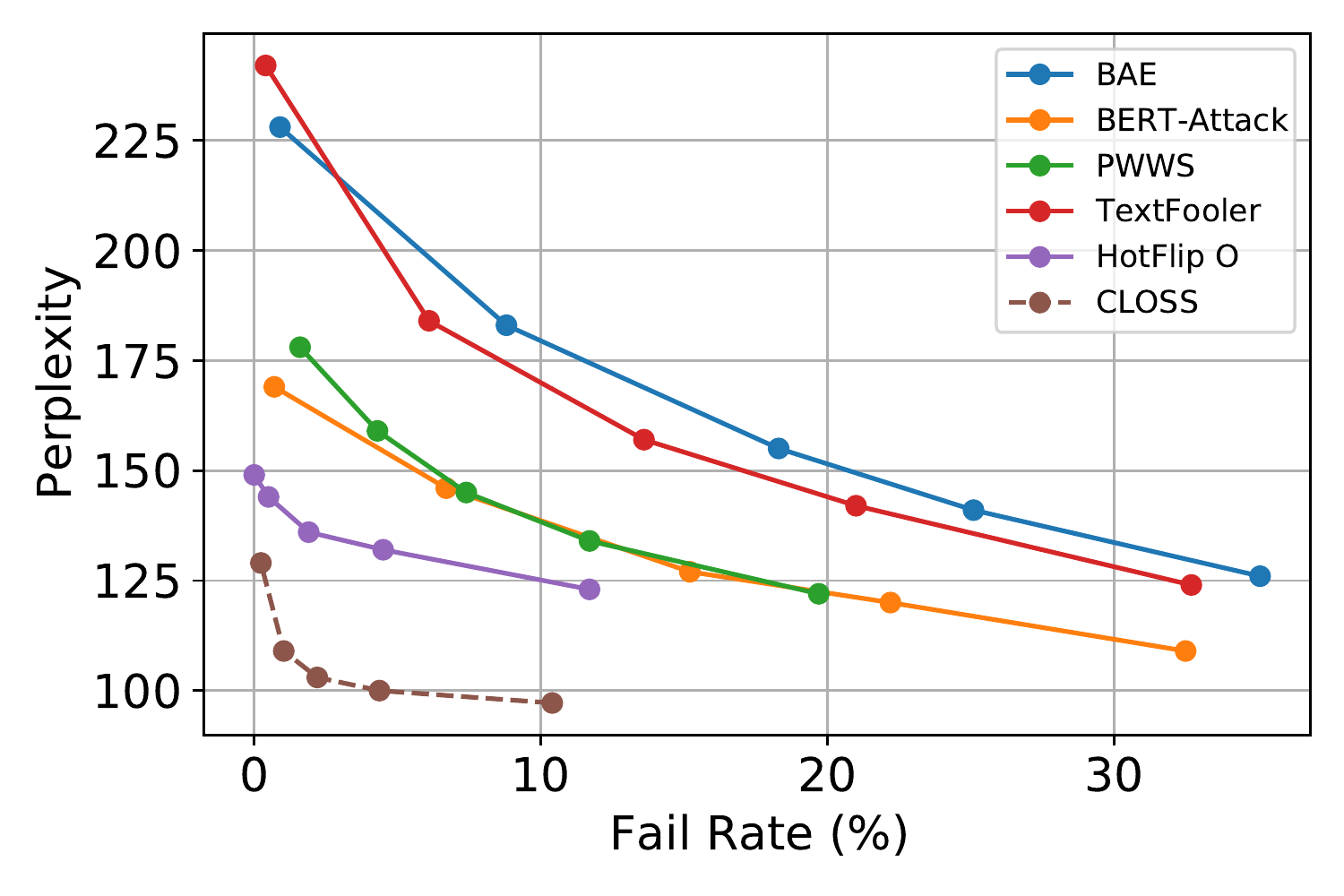} & \includegraphics[scale=0.48]{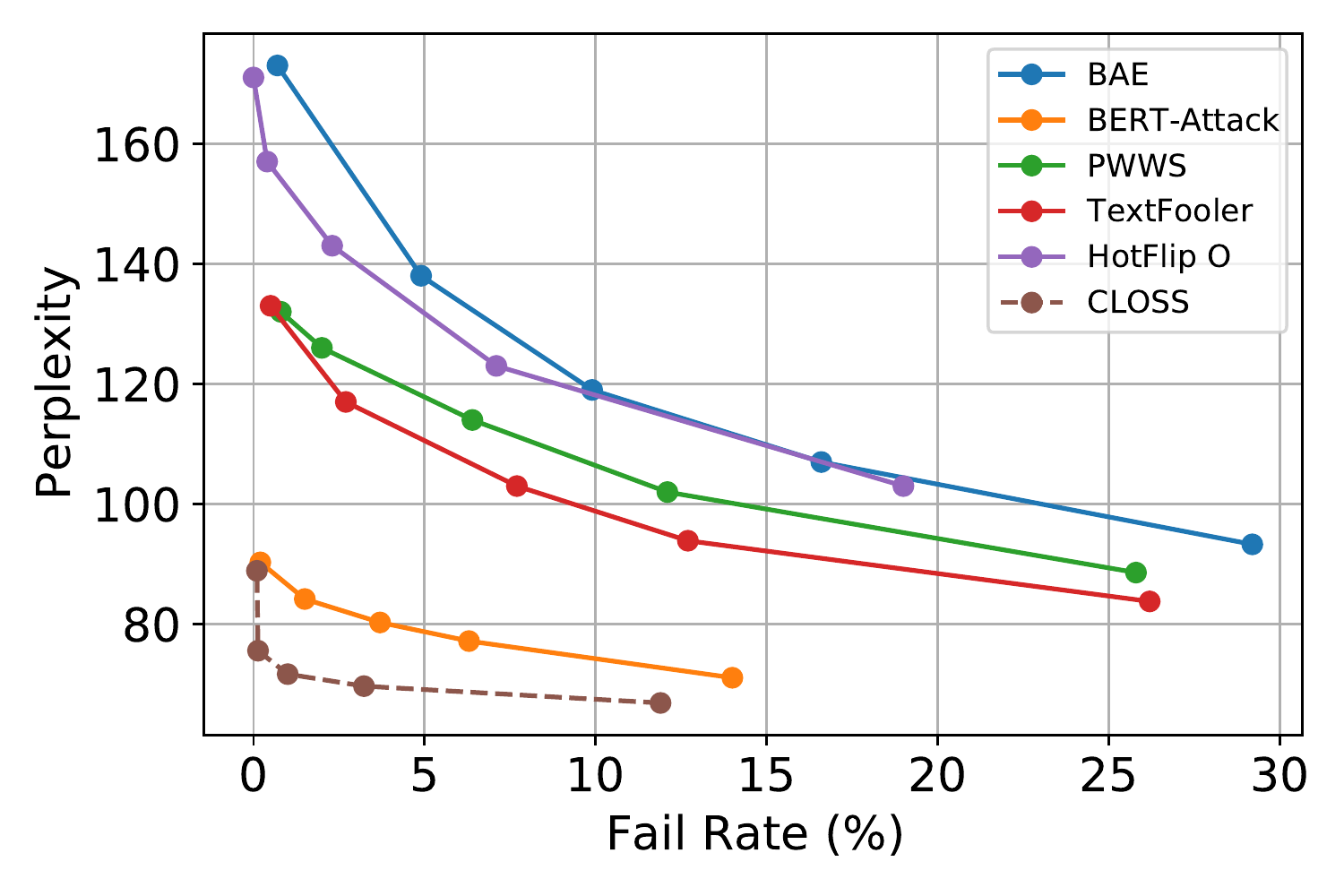} \\
(a) BERT IMDB & (b) RoBERTa IMDB \\
\includegraphics[scale=0.48]{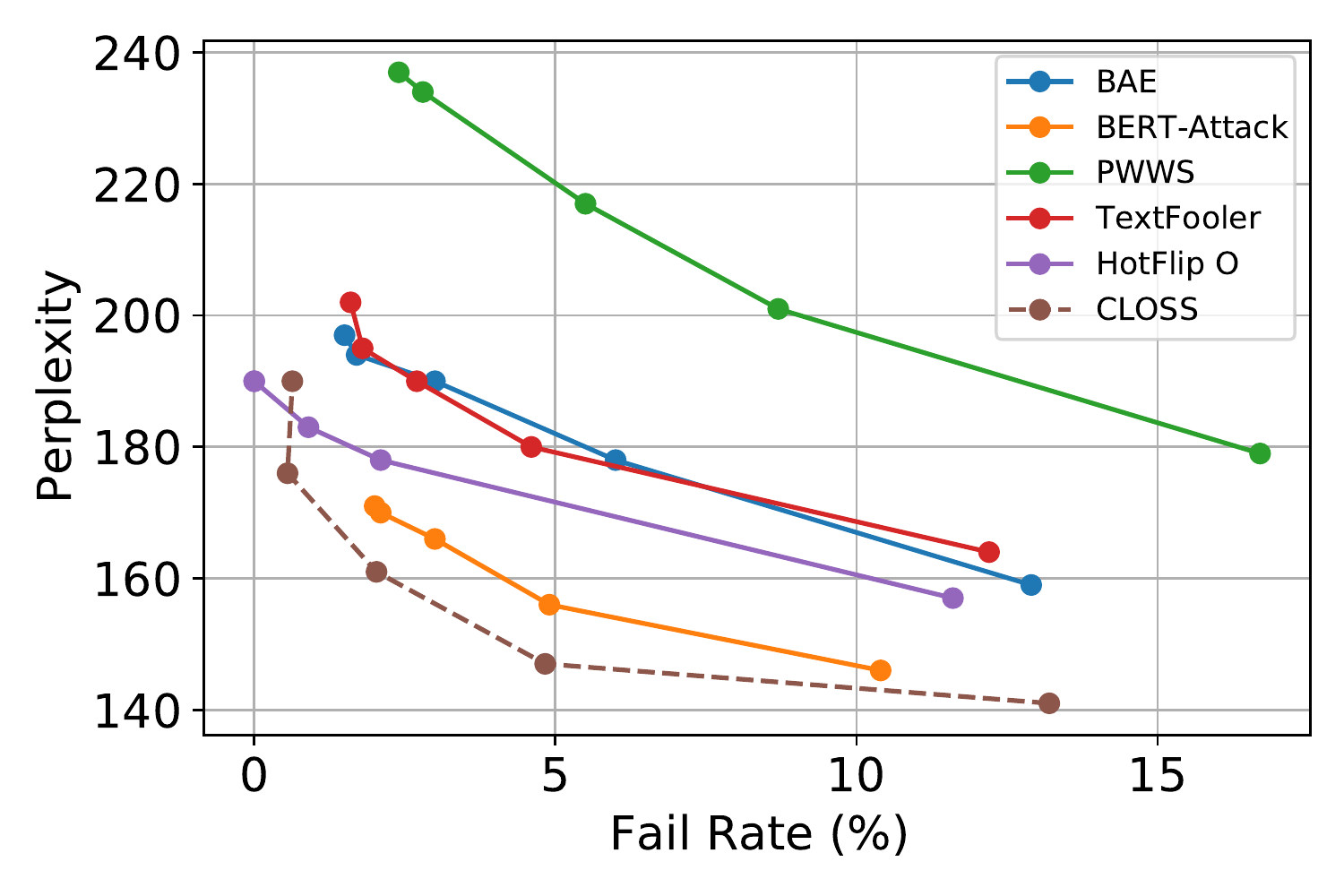} & \includegraphics[scale=0.48]{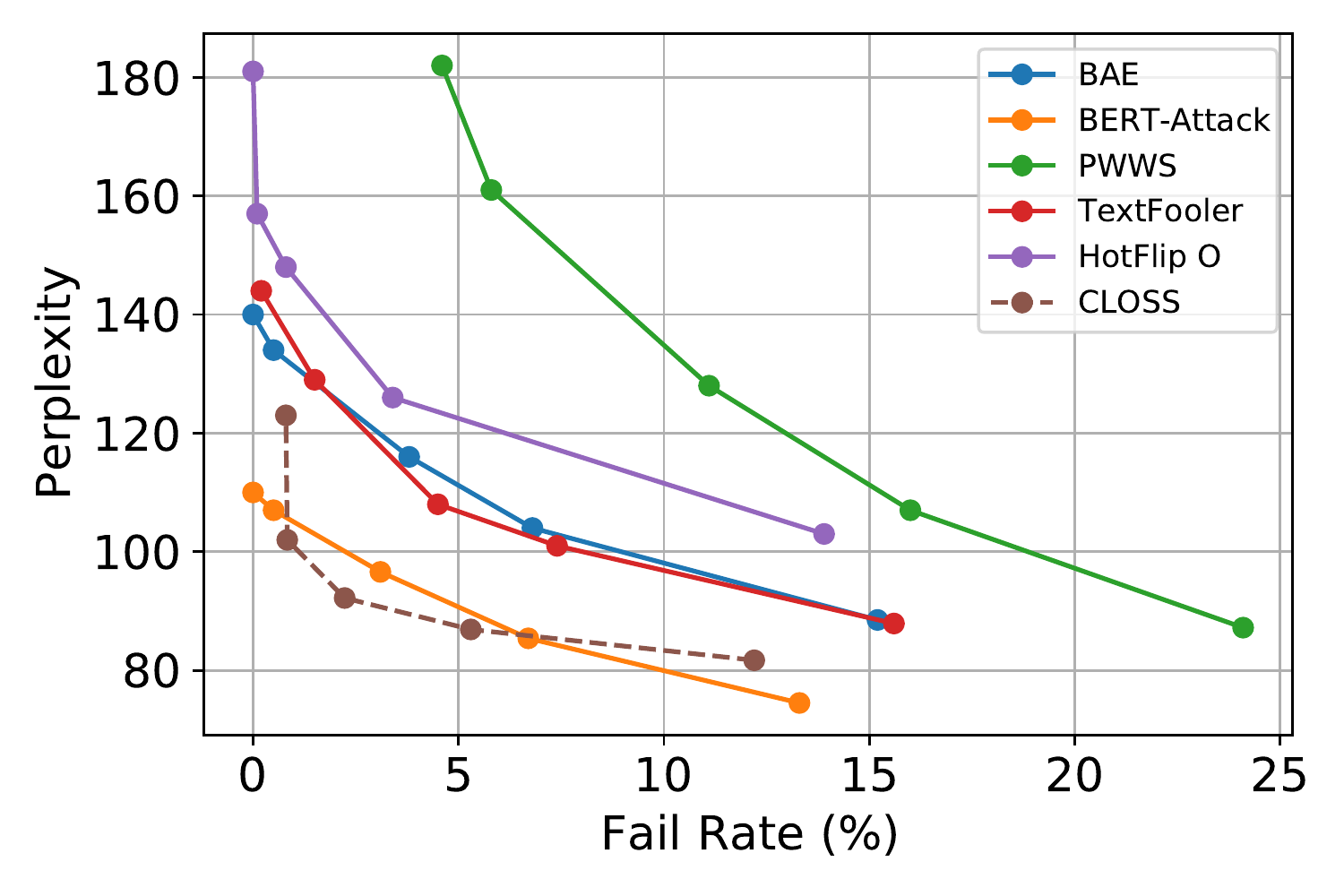} \\
(c) BERT QNLI & (d) RoBERTa QNLI\\
\end{tabular}
\caption{Plots of perplexity against failure rate as the maximum allowed percent of tokens changed $(C_{max})$ varies. Values for $C_{max}$ are 10\%, 15\%, 20\%, 30\% and 50\%. CLOSS values are averaged over three runs.}
\label{figure:differentpercentchanges}
\end{figure*}

\subsection{Impact of Varying $C_{max}$}
\bluecomment{We considering different $C_{max}$ values including 0.1, 0.15, 0.2, 0.3 and 0.5. 
Figure~\ref{figure:differentpercentchanges} plots the perplexity against failure rate for different values of $C_{max}$\footnote{ HotFlip D is excluded due to its poor performance and to preserve the graph scaling for the rest of the methods.}. Increasing $C_{max}$ allows methods to change more input tokens, reducing their failure rates. However, higher $C_{max}$ also leads to greater distortion of the input, raising the perplexity. Thus, methods with better perplexity/fail rate tradeoffs have curves that fall closer to the lower-left corner of the plots. In this regard, CLOSS has the best performance on all comparisons, except for against BERT-Attack on RoBERTa QNLI, where the two methods appear comparable.}


\subsection{Ablation studies}
\label{sec:ablation}
We consider three ablated versions of CLOSS.\\
\textbf{CLOSS-EO} removes the optimization step and instead generates potential substitutions by feeding the original input into the default pre-trained language model associated with model $M$. \\
\textbf{CLOSS-RTL} skips retraining the language modeling head and uses the language modeling head of the pretrained language model. As a result, the language modeling head for this ablation has a different latent space compared to the fine-tuned encoder of classifier $M$. \\
\textbf{CLOSS-SV} removes the Shapley value estimates of each substitution's impact on classification. Instead, we priority substitutions during beam search based on the token saliency (Eq.~\ref{importancescore}).



\begin{table*}[h]
\centering
\begin{tabular}{l | l l l | l l l| l l l | lll}
& \multicolumn{6}{c|}{IMDB} & \multicolumn{6}{|c}{QNLI} \\ 
\hline
& \multicolumn{3}{c}{RoBERTa} &
\multicolumn{3}{c|}{BERT} & \multicolumn{3}{c}{RoBERTa} & \multicolumn{3}{c}{BERT}\\
Method & \%F & \%C & P  & \%F & \%C & P &\%F & \%C & P & \%F & \%C & P \\
\hline
CLOSS & \textbf{4.2} & \textbf{3.13} & 72.4 & \textbf{4.1} & \textbf{2.76} & 98.9 & \textbf{5.1} & \textbf{3.33} & 92.4 & \textbf{3.5} & \textbf{3.31} & 143 \\
CLOSS-SV & 9.4 & 5.73 & 84.5 & 11.6 & 5.06 & 116 & 7.3 & 5.13 & 108 & 6.4 & 5.05 & 159\\
CLOSS-EO & 7.3 & 3.25 & \textbf{63.3} & 8.4 & 3.17 & \textbf{94.9} & 7.2 & 3.51 & \textbf{72.2} & 6.1 & 3.51 & \textbf{122} \\
CLOSS-RTL & 5.5 & 3.2 & 73.7 & 7.5 & 2.9 & 102 & 7.9 & 3.7 & 94.7 & 5.7 & 3.4 & 136\\
\end{tabular}
\caption{
Ablation results on IMDB and QNLI. Values are averaged over three runs.
}
\label{table:finalablation}
\end{table*}

\begin{figure*}[ht!]
\begin{tabular}{ccc}
\includegraphics[scale=0.36]{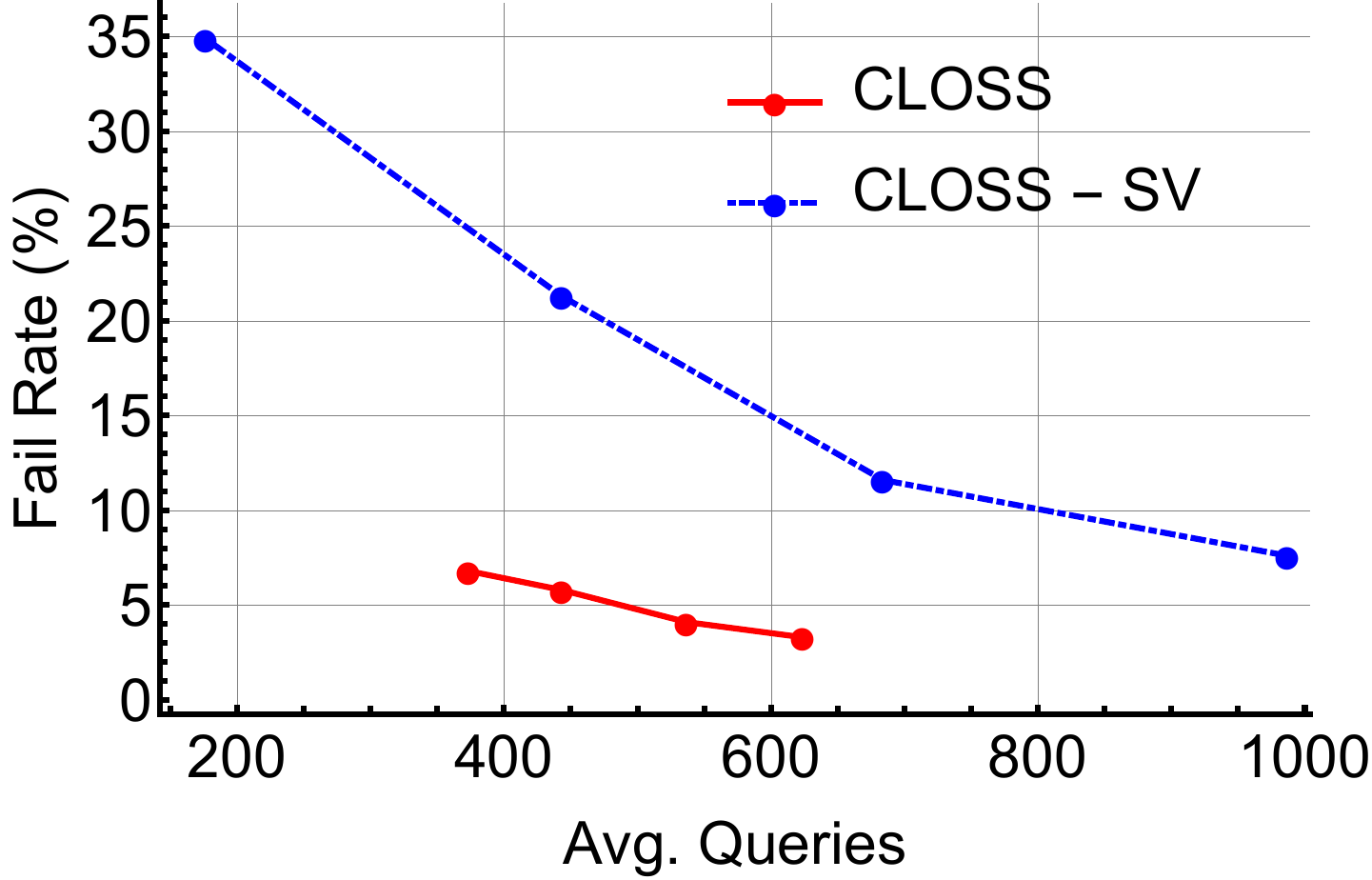} & 
\includegraphics[scale=0.24]{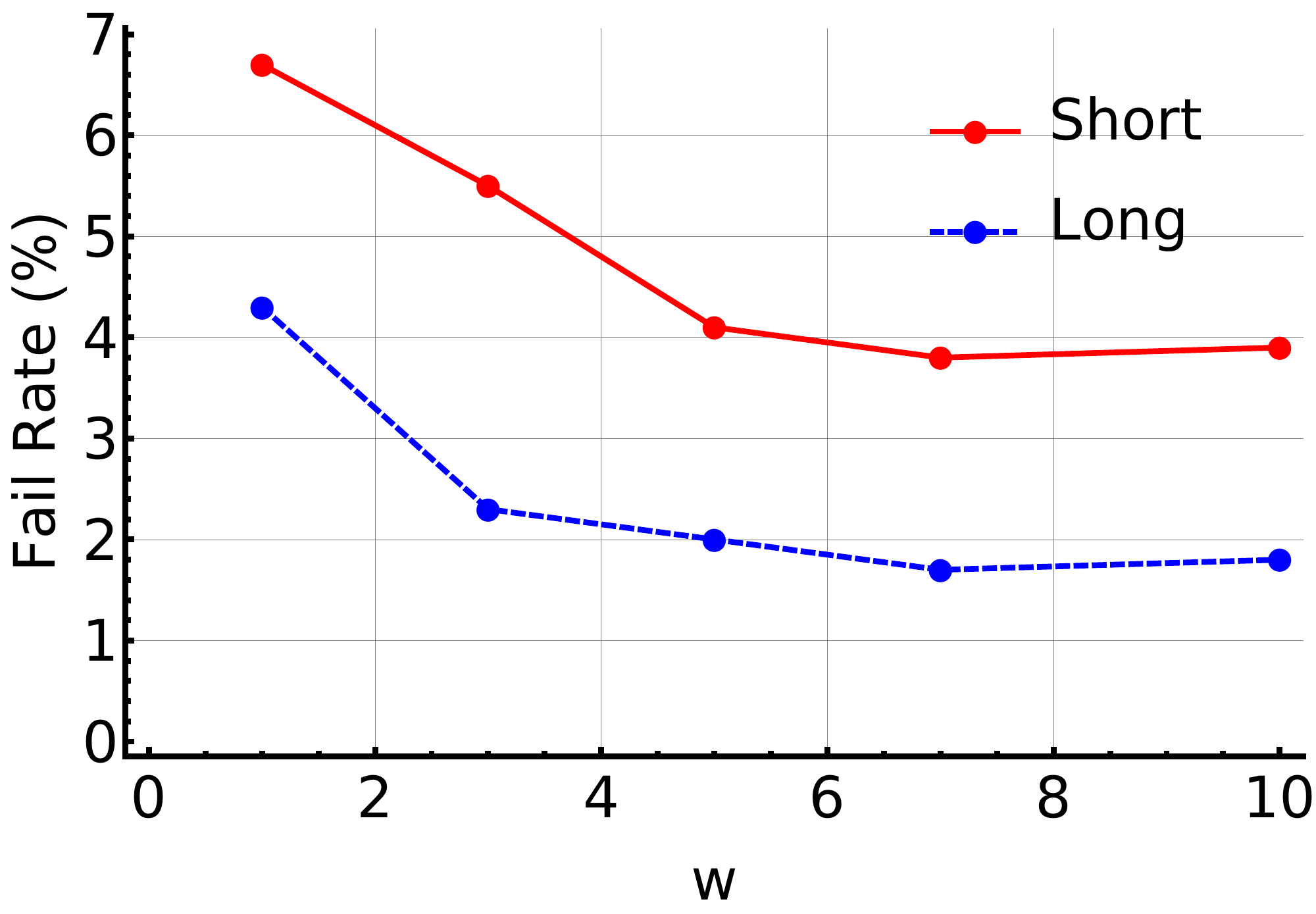}&
\includegraphics[scale=0.24]{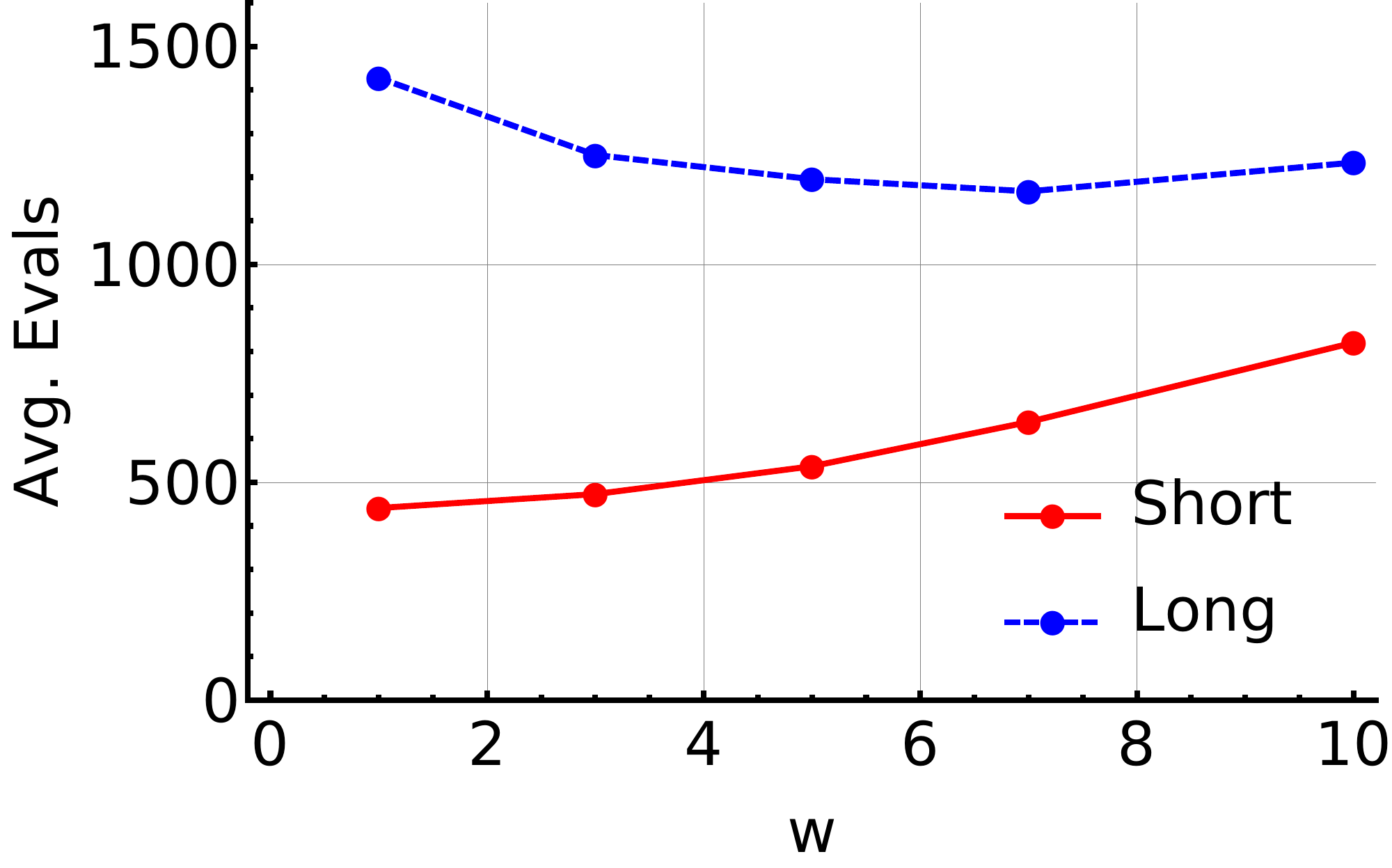}
\\
(a) & (b) & (c) \\
\end{tabular}
\caption{(a) Plot of failure rate of CLOSS and CLOSS-SV as a function of the number of model queries for short IMDB. Beam width ranges from 5 to 20 for both approaches. (b) Plot of failure rate of CLOSS in flipping BERT's prediction as a function of $w$ for both short and long IMDB. (c) Plot of number of BERT model queries used by CLOSS as a function of $w$ for both short and long IMDB.}
\label{figure:threeplots}
\end{figure*}
We compare the performance of CLOSS with its ablations in Table~\ref{table:finalablation}. Here we omit BLEU score because it strongly correlates with \%C in Tables~\ref{table:finalimdbbaselines},  

\paragraph{Effect of embedding optimization.} By removing the optimization step, CLOSS-EO has significantly more failures, but lower perplexity. This is not surprising because optimizing the embedding increases the chance to flip the prediction but carries the risk of producing ``unnatural'' embeddings that lie outside the space of texts previously observed by the language model. This also suggests that CLOSS-EO can be a good candidate for scenarios where "naturalness" of the text is critical. 
\paragraph{Effect of retraining language modeling head.}It is interesting to note that CLOSS-RTL has comparable perplexity to CLOSS, but a higher failure rate. We believe this is because the retrained language modeling head can generate tokens that better match the data distribution of IMDB and QNLI (but not of English text in general), i.e., the distribution of tokens to which the classifier $M$ is sensitive.
\paragraph{Effect of Shapley values.} By removing Shapley value estimates, CLOSS-SV sees substantial degradations in all measures, suggesting critical importance of this step to our method.
\subsection{Computational Considerations}
The estimation of Shapley value for CLOSS incurs a substantial cost in terms of number of queries to the given model. Indeed, the number of queries used by CLOSS can be significantly higher than some baselines\footnote{We report the average number of model queries used by CLOSS and the baselines in Table~\ref{table:baselineQtable} in the Appendix. In practice, CLOSS, implemented without parallelization, can generate counterfactuals for typical inputs in seconds.}. This section takes a closer look at the computational trade-offs surrounding Shapley value estimations.
\paragraph{Shapley or not.} Given the computational cost of estimating Shapley values, would a more thorough search (e.g., using larger beam width) remove the need for computing Shapley values? To explore this question, we consider different beam width $b=5,10,15$ and $20$, and plot the resulting failure rates of CLOSS and CLOSS-SV against the number of queries to the model for short IMDB and BERT in Figure~\ref{figure:threeplots} (a).\footnote{Figures for RoBERTa and QNLI are similar, thus omitted.} The figure shows that even with larger beam width and higher number of queries, the performance of CLOSS-SV still trails behind CLOSS. It also shows that the Shapley value guided search reduces CLOSS's sensitivity to the beam width $b$ both in terms of failure rate and the number of queries needed.  

\paragraph{Precision of Shapley.} The Shapley value is estimated via sampling and the sampling rate is controlled by the parameter $w$. Intuitively, larger $w$ leads to more accurate Shapley values, but incurs higher computation cost. We explore how sensitive our method is to the parameter $w$ in Figure~\ref{figure:threeplots}(b\&c). 

Specifically, Figure~\ref{figure:threeplots}(b) plot the failure rate of CLOSS in flipping BERT's prediction for both short and long IMDB.  We see that as long as $w$ is reasonably large ($\geq$5), the performance is fairly robust. Note that other measures like perplexity and BLEU score show similar trends, which are shown in Figure \ref{figure:thingsvssvs} in the Appendix.  

Figure~\ref{figure:threeplots}(c), on the other hand, plots the average number of queries to model $M$ required with $w$ from 1 to 10. We see an interesting phenomenon for long IMDB where increased $w$ actually leads to decreased number of queries. This may appear counter-intuitive at first sight, it actually demonstrates the power of good Shapley value estimates in speeding up the search. This phenomenon, however, was not observed for the short IMDB data, likely due to the substantially smaller search space thanks to the shorter input length.

\section{Human and Qualitative Evaluation}

\paragraph{Human evaluations.}
For human evaluations, we choose to compare CLOSS against Bert-Attack and HotFlip O, the two baselines performing the best in perplexity and flip rate respectively. 


We randomly selected 100 original texts from IMDB for evaluation with the restriction that all three methods must successfully flip the classification changing 15\% or less of the original tokens. Additionally, we exclude texts with more than 50 tokens to ease the burden on evaluators. Using the BERT classifier, We apply BERT-Attack, CLOSS and HotFlip to generate counterfactuals for each input. Eight human evaluators are each assigned 25 original texts and asked to rank (ties allowed) the three counterfactuals in order of grammatical correctness. Each input is evaluated by two evaluators, the inter-evaluator agreement per pairwise comparison is 75.4\%.
Human evaluators ranked CLOSS competitively with BERT-Attack and HotFlip, assigning average ranks of 1.54 to CLOSS, 1.68 to Bert-Attack and 2.50 to HotFlip. The difference between CLOSS and HotFlip is statistically significant (one-sided sign test, $p$-value$<0.0001$). 

\paragraph{Qualitative analysis of generated text.}
Inspecting the generated coutnerfactuals, we observe some interesting patterns, summarized below. 
See Appendix (Tables~\ref{table:imdbquals} and \ref{table:qnliquals}) for specific examples.

For the IMDB dataset, CLOSS often changes one or two sentiment words while the rest of the input still supports the original prediction. This suggests that the model may be triggered by a few sentiment words, ignoring most input. Identifying such critical substitutions will allow us to inspect the patterns of these “triggers” to reveal the weakness of the classifier. We also observe that when the model misclassifies, it often takes little change to correct the model, which helps debug the  mistake.

Sometimes CLOSS introduces synergistic changes where each change's capacity to influence the classification seems contingent on the other. Finally, CLOSS sometimes distorts sentiment-phrases into non-words to remove their impact on classification, possibly making up for the lack of ability to remove words.

For the QNLI dataset, unsurprisingly, we note that changing from entailment to non-entailment is far easier than the opposite (see Figures~\ref{fig:fracchanges}(c,d) in Appendix), and often requires changing only a few words shared by the Question and Context. Conversely, CLOSS can sometimes change non-entailment to entailment by introducing some shared word(s). This suggests that the model relies heavily on overlapping words to decide entailment.

More detailed analysis can be found in the Appendix, including how CLOSS's changes are distributed among part of speech tags (Figure~\ref{figure:poschanges}) and a failure analysis for CLOSS (A.7, Table~\ref{table:erroranalysis}). 

\section{Conclusion}

We are motivated by how humans use counterfactuals to explain the concept of a class and seek to automatically generate counterfactual text input as a means to understand a deep NLP model and its definition of class. 
We assume full white-box access to the given model and perform optimization in the latent space to maximize the probability of predicting a target class. We then map from the optimized latent representation to candidate token substitutions using a language model. A key novelty of CLOSS is using Shapley values to estimate the potential of a token substitution in changing the model's prediction when used in combination with other substitutions. The Shapley value is then used to guide a breadth-first beam search to generate the final counterfactual. Through both automatic and human evaluations, we show that CLOSS achieves highly competitive performance both in terms of the success rate of generating counterfactuals as well as the quality of the generated counterfactuals. 

Our approach has several limitations. As a white-box approach, we require full access to the model, which can be restrictive in practical applications. Our approach currently only considers substitutions, excluding deletions and insertions. Finally, our method is only applicable to models that are based on pre-trained language models. Future work will adapt CLOSS to adversarial and black box settings. We also hope to improve the efficiency of CLOSS via more efficient Shapley value estimation \cite{chen2018lshapley,jia2020efficient}.

\section*{Acknowledgements}
This work was partially supported by DARPA under grant N66001-17-2-4030.

\bibliography{anthology,custom}
\bibliographystyle{acl_natbib}

\appendix

\section{Appendix}
\label{sec:appendix}

\subsection{Optimized HotFlip.} For each candidate in original HotFlip's beam search, we score every possible single-token substitution by using gradients to estimate the substitution's impact on the classification. The score of a candidate counterfactual is the sum of the scores of each individual substitution introduced by the candidate.
    
These scores form a surrogate value function, which the beam search aims to maximize. At each step of the beam search, we can generate the successors (children) for each current beam members (parent) by applying a single substitution to any location in the parent text.
    
In our optimized HotFlip, we change the search procedure to promote diversity in the beam search by requiring every child generated off a common parent modify distinct locations in the text. We observe that this small modification substantially boost the performance HotFlip's performance. We also increase the beam size from the 10 suggested by \citet{ebrahimi2018hotflip} to 100. Note that the original HotFlip's parameters are designed for character-level modification, which has a substantially smaller space of possible substitutions for each location. This might explain the the poor performance of HotFlip D, and the need to modify the search procedure for token-level generation.

\subsection{Average Number of Queries}
\label{sec:appendix-query}
\begin{table*}[h]
\begin{tabular}{l | cc | cc }
& \multicolumn{2}{c}{IMDB} & \multicolumn{2}{c}{QNLI}\\
Method & RoBERTa & BERT & RoBERTa & BERT \\
\hline
CLOSS & 558$|$485 & 537$|$457 & 455$|$405 & 424$|$389 \\
\hline
HotFlip D & 65.3$|$47.5  & 51.7$|$38.0 & 28.7$|$22.4 & 27.0$|$19.9 \\
HotFlip O & 305$|$265 & 234$|$205 & 90.0$|$82.3 & 71.0$|$66.7 \\
\hline
BAE  & 142$|$111 & 138$|$110 & 94.7$|$81.7 & 97.1$|$85.7 \\
PWWS  & 483$|$466 & 475$|$455 & 246$|$236 & 238$|$236 \\
Textfooler  & 259$|$190 & 298$|$171 & 144$|$102 & 127$|$104 \\
\hline

BAE-U & 424$|$285 & 494$|$252 & 172$|$148 & 180$|$142 \\
BERT-Attack-U & 277$|$231 & 451$|$262 & 172$|$149 & 182$|$140 \\
PWWS-U & 575$|$561 & 568$|$549 & 299$|$290 & 290$|$287 \\
Textfooler-U & 358$|$272 & 426$|$259 & 181$|$152 & 179$|$145 \\
\end{tabular}
\caption{
Average queries per sample for CLOSS and  baselines.
}
\label{table:baselineQtable}
\end{table*}

In Table~\ref{table:baselineQtable}, we present the average number of queries to the given model for CLOSS and baseline methods.  We show two numbers per cell, where the first number is the average query number of all attempts (success or fail). The second number is the average number of queries for successful trials that modify 15\% or less of input tokens. 


\begin{figure*}[ht!]
\begin{tabular}{ccc}
\includegraphics[scale=0.24]{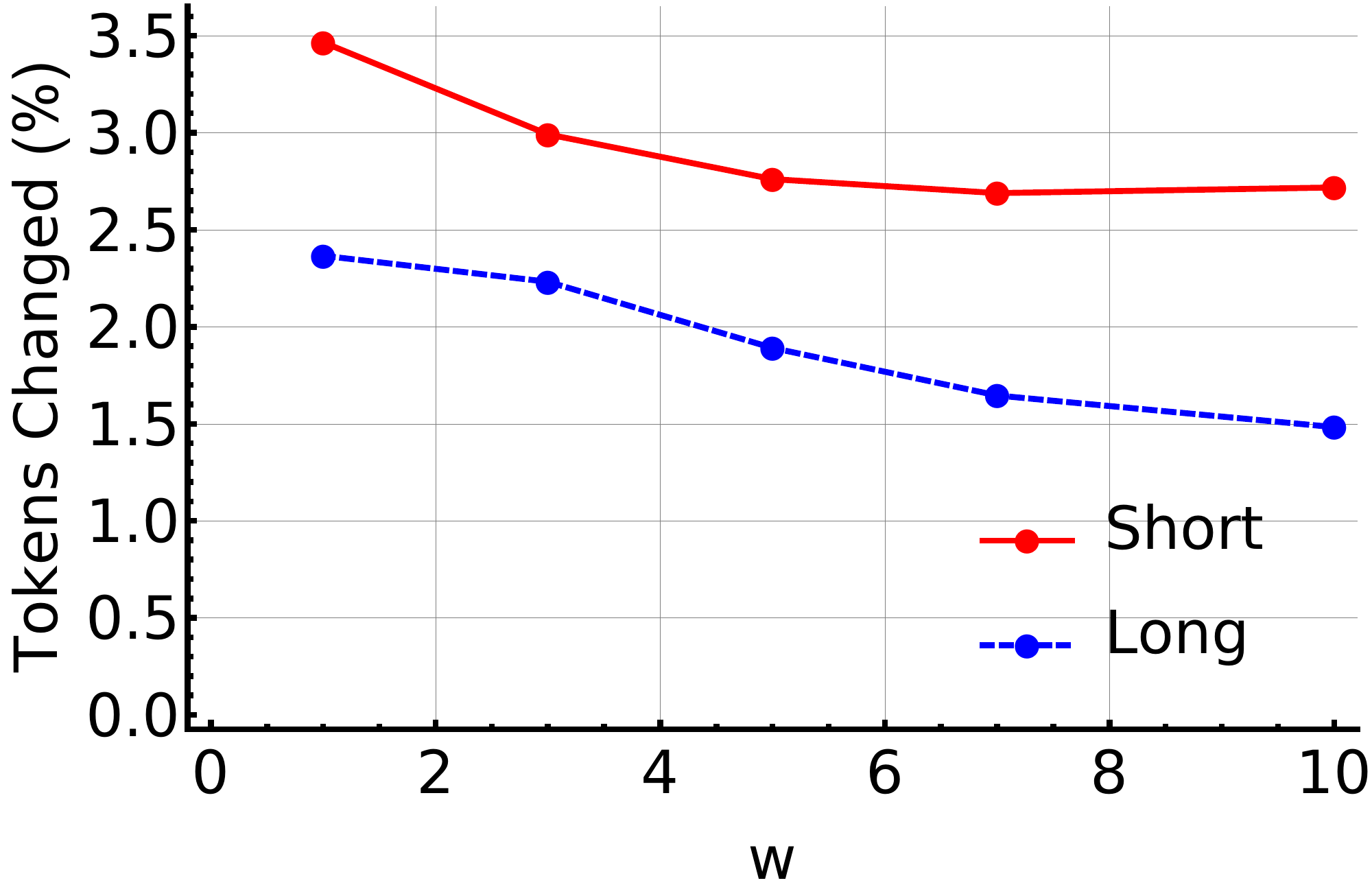} &
\includegraphics[scale=0.24]{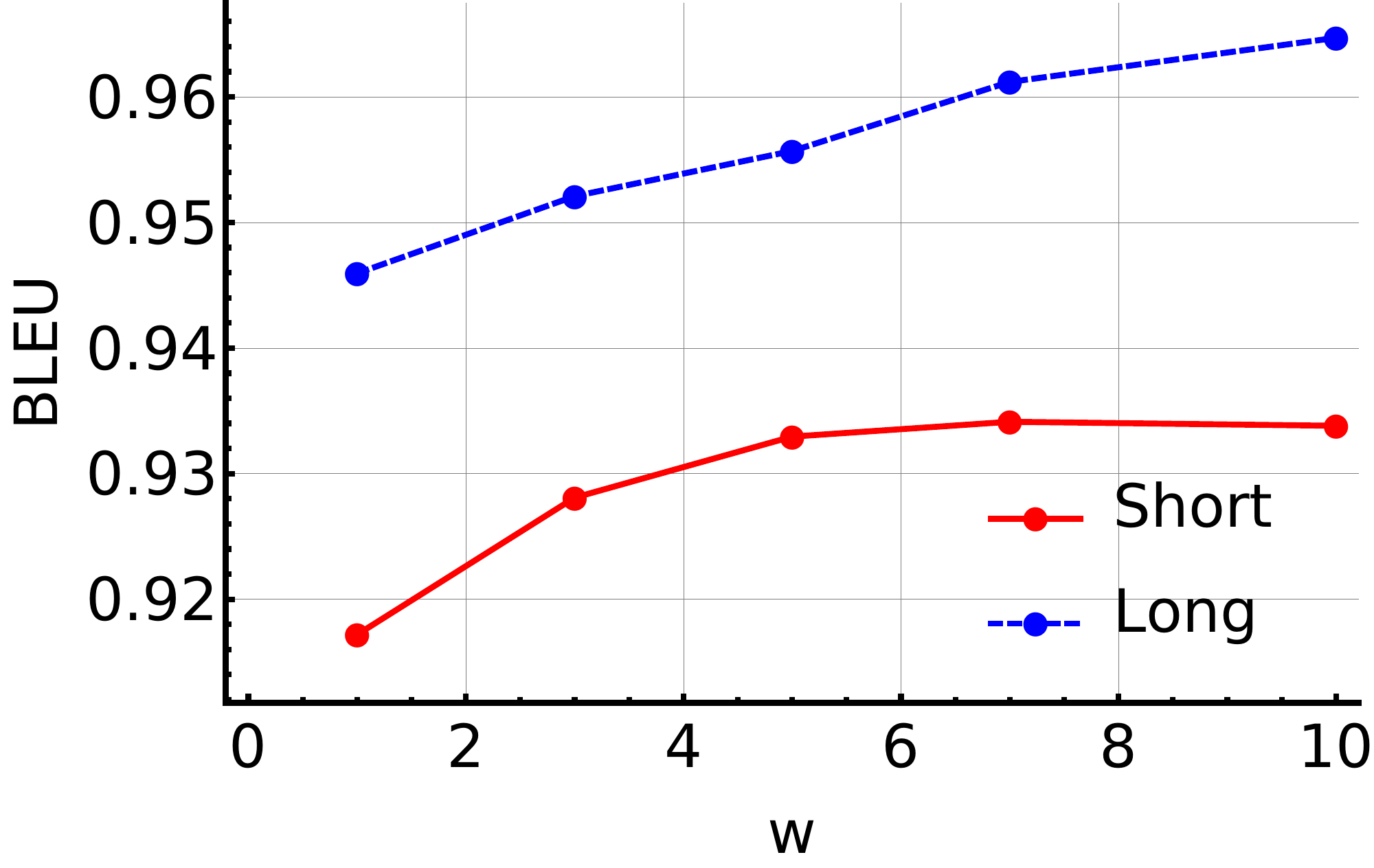} & 
\includegraphics[scale=0.24]{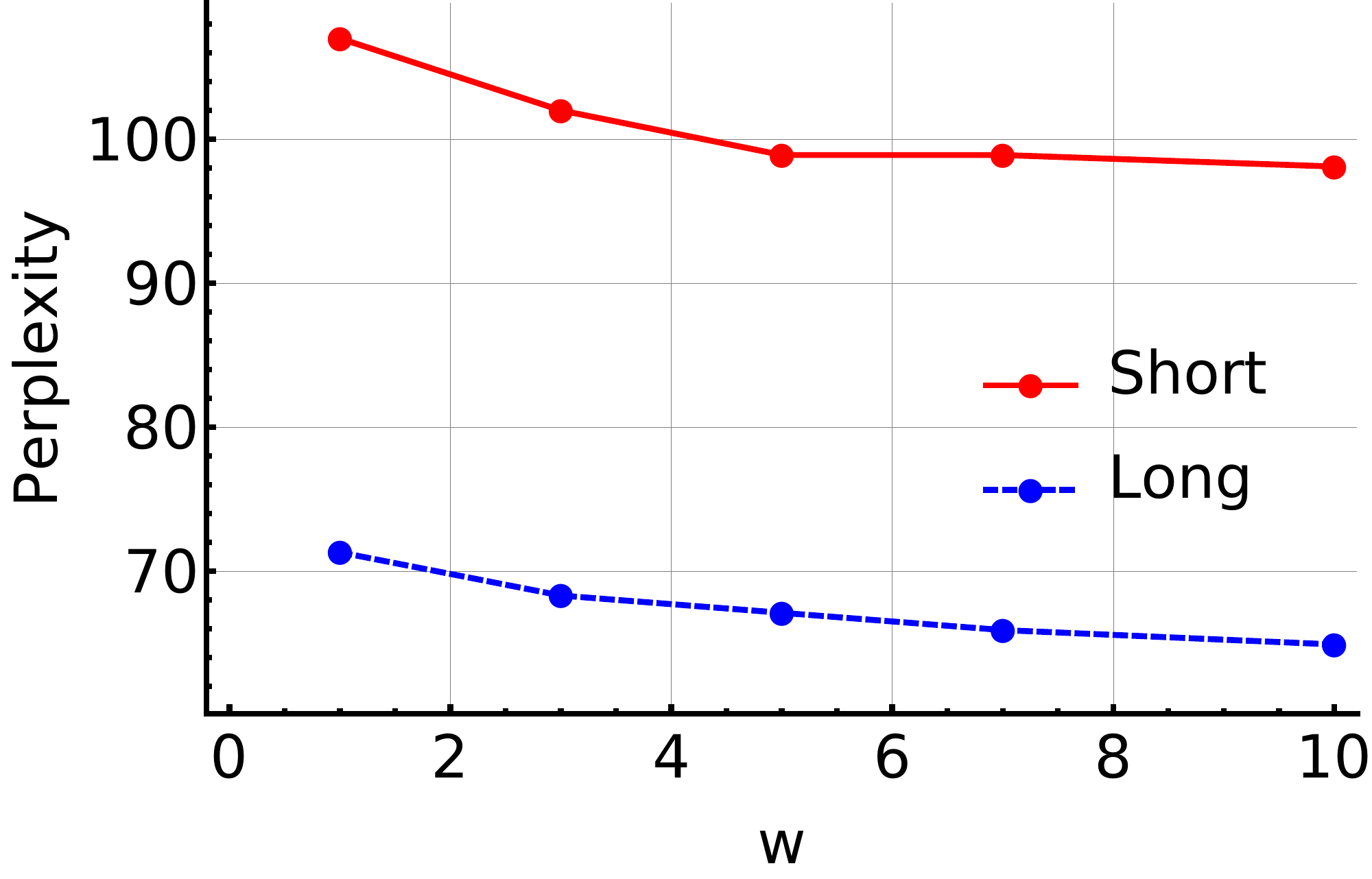}\\
(a) & (b) & (c) \\
\end{tabular}
\caption{(a) Plot of average percent tokens changed by CLOSS as a function of $w$ for both short and long IMDB. (b) Plot of CLOSS average BLEU score as a function of $w$ for both short and long IMDB. (c) Plot of CLOSS average perplexity as a function of $w$ for both short and long IMDB.}
\label{figure:thingsvssvs}
\end{figure*}
\subsection{Performance when varying $w$}
In Figure~\ref{figure:thingsvssvs}, we plot the other performance measures including \%C, BLEU and Perlexity, as a function of parameter $w$.

\subsection{Qualitative Analysis}

In tables \ref{table:imdbquals} and \ref{table:qnliquals}, we present examples of counterfactuals generated by CLOSS that highlight interesting patterns we notice.

\begin{table*}[]
    \centering
    \begin{tabular}{p{0.2\linewidth} | p{0.74\linewidth}}
        Description & Text \\
        \hline
        (a) We can flip the class by changing a small fraction of the sentiment regions.
        &\textbf{Old:}  Ruth Gordon is one of the more sympathetic killers that Columbo has ever had to deal with. And, the plot is ingenious all the way around. This is one of the \textcolor{red}{best} Columbo episodes ever. Mariette Hartley and G. D. Spradlin are excellent in their supporting roles. And Peter Falk delivers a little something extra in his scenes with Gordon.$\newline$
           
        \textbf{New:}  Ruth Gordon is one of the more sympathetic killers that Columbo has ever had to deal with. And, the plot is ingenious all the way around. This is one of the \textcolor{red}{worse} Columbo episodes ever. Mariette Hartley and G. D. Spradlin are excellent in their supporting roles. And Peter Falk delivers a little something extra in his scenes with Gordon.$\newline$ $\newline$
           
        \textbf{Old:}  ruth gordon is one of the more sympathetic killers that columbo has ever had to deal with. and, the plot is ingenious all the way around. this is one of the \textcolor{red}{best} columbo episodes ever. mariette hartley and g. d. spradlin are excellent in their supporting roles. and peter falk delivers a little something extra in his scenes with gordon.$\newline$
    
        \textbf{New:}  ruth gordon is one of the more sympathetic killers that columbo has ever had to deal with. and, the plot is ingenious all the way around. this is one of the \textcolor{red}{worst} columbo episodes ever. mariette hartley and g. d. spradlin are excellent in their supporting roles. and peter falk delivers a little something extra in his scenes with gordon. \\
        
        \hline
        
        (b) We sometimes see synergistic changes where each change's capacity to influence the classification seems contingent on the other.
        &\textbf{Old:}  \textcolor{red}{Excellent} \textcolor{red}{documentary} that still manages to shock and enlighten. Unfortunately, times haven't changed much since this was made and it is thus an important piece for all freedom-conscious Americans to see.$\newline$
        
        \textbf{New:}  \textcolor{red}{Very} \textcolor{red}{pathetic} that still manages to shock and enlighten. Unfortunately, times haven't changed much since this was made and it is thus an important piece for all freedom-conscious Americans to see.$\newline$ $\newline$

        \textbf{Old:} I love all his work but this \textcolor{green}{looks} like \textcolor{green}{nothing}.. sorry.. This looks more like a "David Lynch copycat". I think people like it only because "it's from David Lynch".$\newline$ $\newline$
        \textbf{New:}  I love all his work but this \textcolor{green}{hits} like \textcolor{green}{everything}.. sorry.. This looks more like a "David Lynch copycat". I think people like it only because "it's from David Lynch". \\
        


        \hline
    \end{tabular}
\end{table*}

\begin{table*}[]
    \centering
    \begin{tabular}{p{0.2\linewidth} | p{0.74\linewidth}}
        Description & Text \\
        \hline
        
        (c) RoBERTa incorrectly classified text as positive. Flipping to negative requires little changes.
        
        &\textbf{Old:}  Some \textcolor{red}{good} movies keep you in front of the TV, and you are dying to see the result.$\newline$This movie does not have highs and lows. It simply describes a young girl's family life in Africa. People come and go, the weather and the background are all the same.$\newline$
        
        \textbf{New:}  Some \textcolor{red}{decent} movies keep you in front of the TV, and you are dying to see the result.$\newline$This movie does not have highs and lows. It simply describes a young girl's family life in Africa. People come and go, the weather and the background are all the same.\\
        
        \hline
        
        (d) BERT classifies as negative. Greater changes required to flip to positive.
        &\textbf{Old:}  some good movies keep you in front of the tv, and you are \textcolor{green}{dying} to see the result. $\newline$ this movie does not \textcolor{green}{have} highs and lows. it simply describes a young girl's family life in africa. people come and go, the weather and the background are all the same.$\newline$
         
        \textbf{New:}  some good movies keep you in front of the tv, and you are \textcolor{green}{loving} to see the result. $\newline$ this movie does not \textcolor{green}{lack} highs and lows. it simply describes a young girl's family life in africa. people come and go, the weather and the background are all the same.\\
        
        \hline 
        
        (e) Sometimes distorts words/grammar; Note how CLOSS removes "I loved this" by convering "loved this" into "lovedoo", thereby removing the original's positive sentiment
        &\textbf{Old:}  I loved \textcolor{red}{this} mini series. Tara Fitzgerald did an incredible job portraying Helen Graham, a beautiful young woman hiding, along with her young son, from a mysterious past. As an anglophile who loves romances... this movie was just my cup of tea and I would recommend it to anyone looking to escape for a few hours into the England of the 1800's. I also must mention that Toby Stephens who portrays the very magnetic Gilbert Markham \textcolor{red}{is} reason enough to watch this \textcolor{red}{wonderful} production.$\newline$
        
        \textbf{New:}  I \textcolor{red}{lovedoo} mini series. Tara Fitzgerald did an incredible job portraying Helen Graham, a beautiful young woman hiding, along with her young son, from a mysterious past. As an anglophile who loves romances... this movie was just my cup of tea and I would recommend it to anyone looking to escape for a few hours into the England of the 1800's. I also must mention that Toby Stephens who portrays the very magnetic Gilbert Markham \textcolor{red}{does} reason enough to watch this \textcolor{red}{dreadful} production.\\
        
        \hline
        
        (f) Non-words can significantly change sentiment classification. "thisecrated" doesn't seem particularly sentiment-related, yet it can flip the classification of this otherwise very positive review.
        &\textbf{Old:}  absolutely fantastic! whatever i say wouldn't do this \textcolor{red}{underrated} movie the justice it deserves. watch it now! fantastic!$\newline$
         
        \textbf{New:}  absolutely fantastic! whatever i say wouldn't do \textcolor{red}{thisecrated} movie the justice it deserves. watch it now! fantastic! \\
        
        \hline

    \end{tabular}
    \caption{Example IMDB counterfactuals generated by CLOSS. Each row demonstrates an interesting pattern of behavior we observed. We use green to highlight words whose changes flip the text to positive and red for changes that flip texts to negative.}
    \label{table:imdbquals}
\end{table*}

\begin{table*}[]
    \centering
    \begin{tabular}{p{0.2\linewidth} | p{0.74\linewidth}}
        Description & Text \\
        \hline
        (a) CLOSS can often flip entilment to non-entailment by changing a word that appears in both the Question and Context.
        &\textbf{Old:} 
        Question: When was Luther's \textcolor{red}{last} sermon? $\newline$
        Context : His \textbf{last} sermon was delivered at Eisleben, his place of birth, on 15 February 1546, three days before his death. $\newline$
        
        \textbf{New:} Question: When was Luther's \textcolor{red}{new} sermon? $\newline$
        Context : His \textbf{last} sermon was delivered at Eisleben, his place of birth, on 15 February 1546, three days before his death. $\newline$ $\newline$

        \textbf{Old:} Question: when was luther's \textcolor{red}{last} sermon? $\newline$
        Context : his \textbf{last} sermon was delivered at eisleben, his place of birth, on 15 february 1546, three days before his death. $\newline$
        
        \textbf{New:} 
        Question: when was luther's \textcolor{red}{traveling} sermon? $\newline$
        Context : his \textbf{last} sermon was delivered at eisleben, his place of birth, on 15 february 1546, three days before his death. \\

        \hline

        (b) CLOSS can sometimes induce entailment by changing a word in the Question (Context) to match one in the Context (Question).
        &\textbf{Old:} 
        Question: Who were the ESPN Deportes \textcolor{green}{commentators} for Super Bowl 50? $\newline$
        Context : On December 28, 2015, ESPN Deportes announced that they had reached an \textbf{agreement} with CBS and the NFL to be the exclusive Spanish-language broadcaster of the game, marking the third dedicated Spanish-language broadcast of the Super Bowl.$\newline$
        
        \textbf{New:} 
        Question: Who were the ESPN Deportes \textcolor{green}{agreements} for Super Bowl 50?$\newline$
        Context : On December 28, 2015, ESPN Deportes announced that they had reached an \textbf{agreement} with CBS and the NFL to be the exclusive Spanish-language broadcaster of the game, marking the third dedicated Spanish-language broadcast of the Super Bowl.\\

        \hline
        (c) If lexical overlap fails, we often need many edits to change non-entailment to entailment.
        &\textbf{Old:} 
        Question: Who was the \textcolor{green}{number} two draft pick for \textcolor{green}{2011}?$\newline$
        Context : This was the first Super \textcolor{green}{Bowl} to feature a \textcolor{green}{quarterback} on both teams who was the \#1 pick in their draft classes.$\newline$
        
        \textbf{New:} 
        Question: Who was the \textcolor{green}{show} two draft pick for \textcolor{green}{Kate}?$\newline$
        Context : This was the first Super \textcolor{green}{half} to feature a \textcolor{green}{Premier} on both teams who was the \#1 pick in their draft classes.\\
        
        \hline
    \end{tabular}
    \caption{Example QNLI counterfactuals generated by CLOSS. Each row demonstrates an interesting pattern of behavior we observe. We use green to highlight words whose changes flip the text to entailment and red for changes that flip texts to non-entailment.}
    \label{table:qnliquals}
\end{table*}


\subsection{Part of Speech Changes}
\label{asect:pos}
In Figure \ref{figure:poschanges}, we show the percentages of total changes that occur in each part of speech type. We split the results by direction of change. Note that for IMDB, CLOSS tends to modify adjectives more when flipping from negative to positive compared to flipping from positive to negative.

When flipping entailment to non-entailment, CLOSS is more likely to modify nouns compared to flipping non-entailment to entailment. This may re

\begin{figure*}[]
    \centering
    \begin{tabular}{c c}
        \includegraphics[scale=0.5]{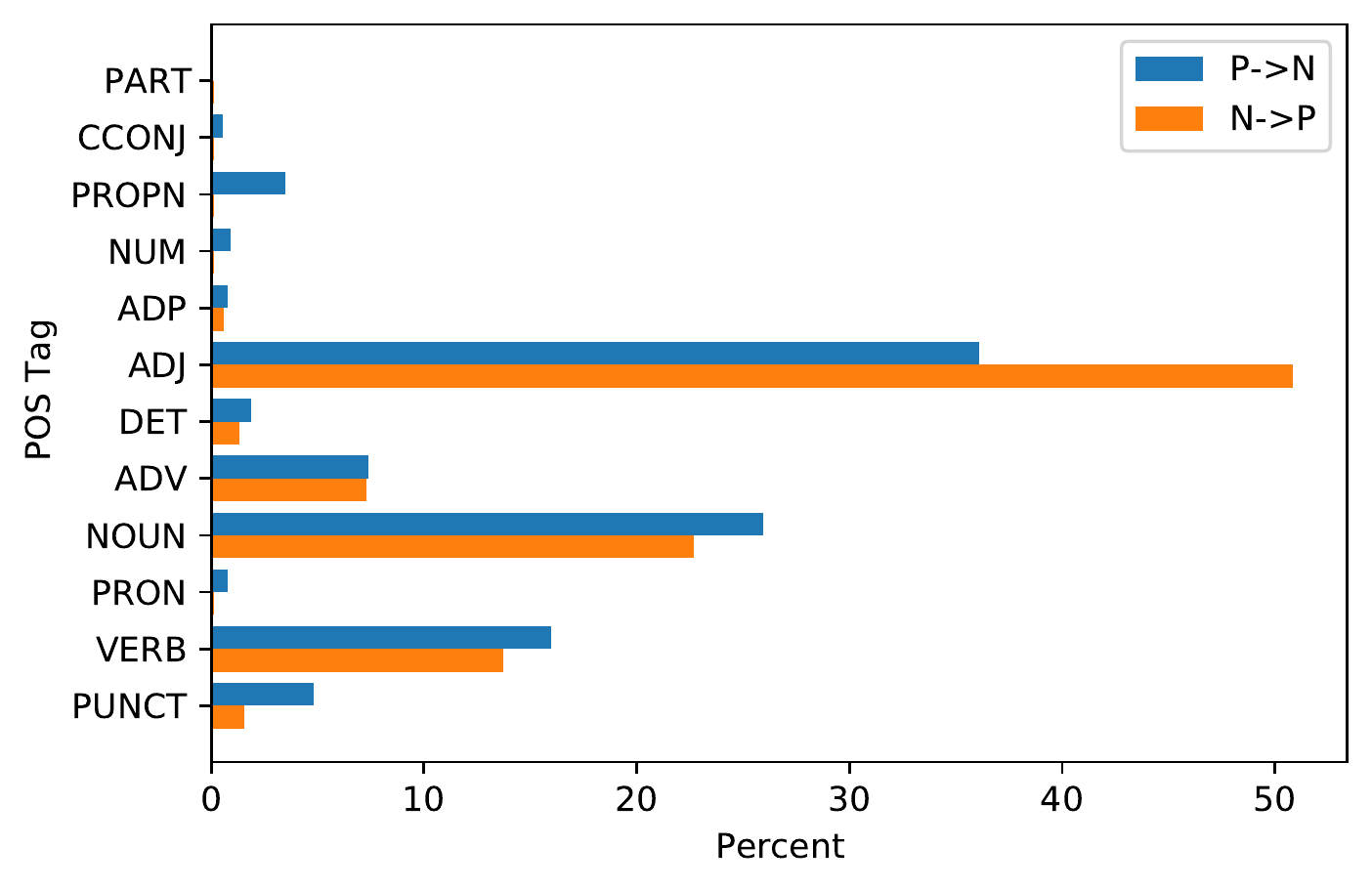} &  \includegraphics[scale=0.5]{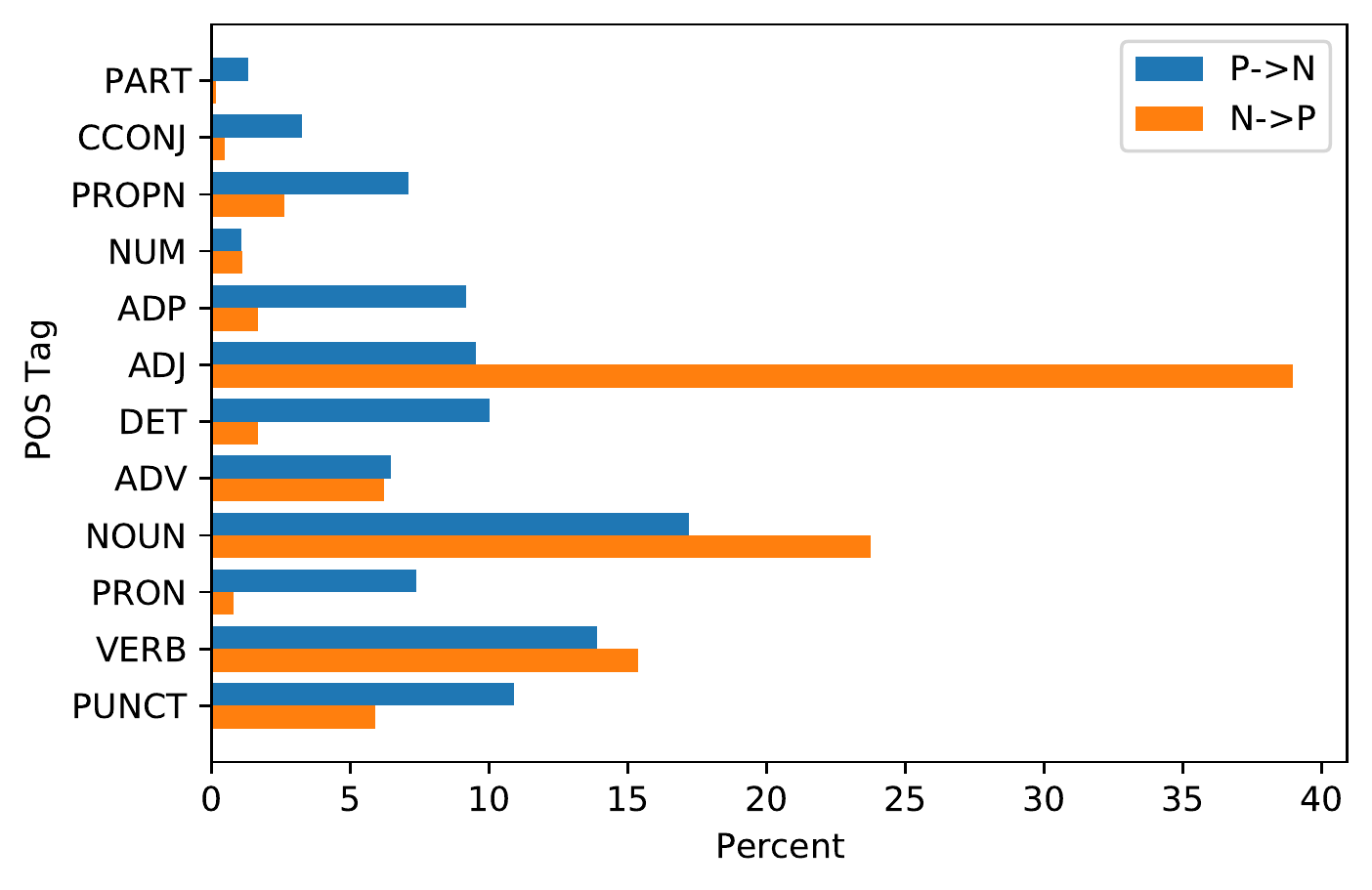}\\
        (a) BERT IMDB & (b) RoBERTa IMDB\\
        \includegraphics[scale=0.5]{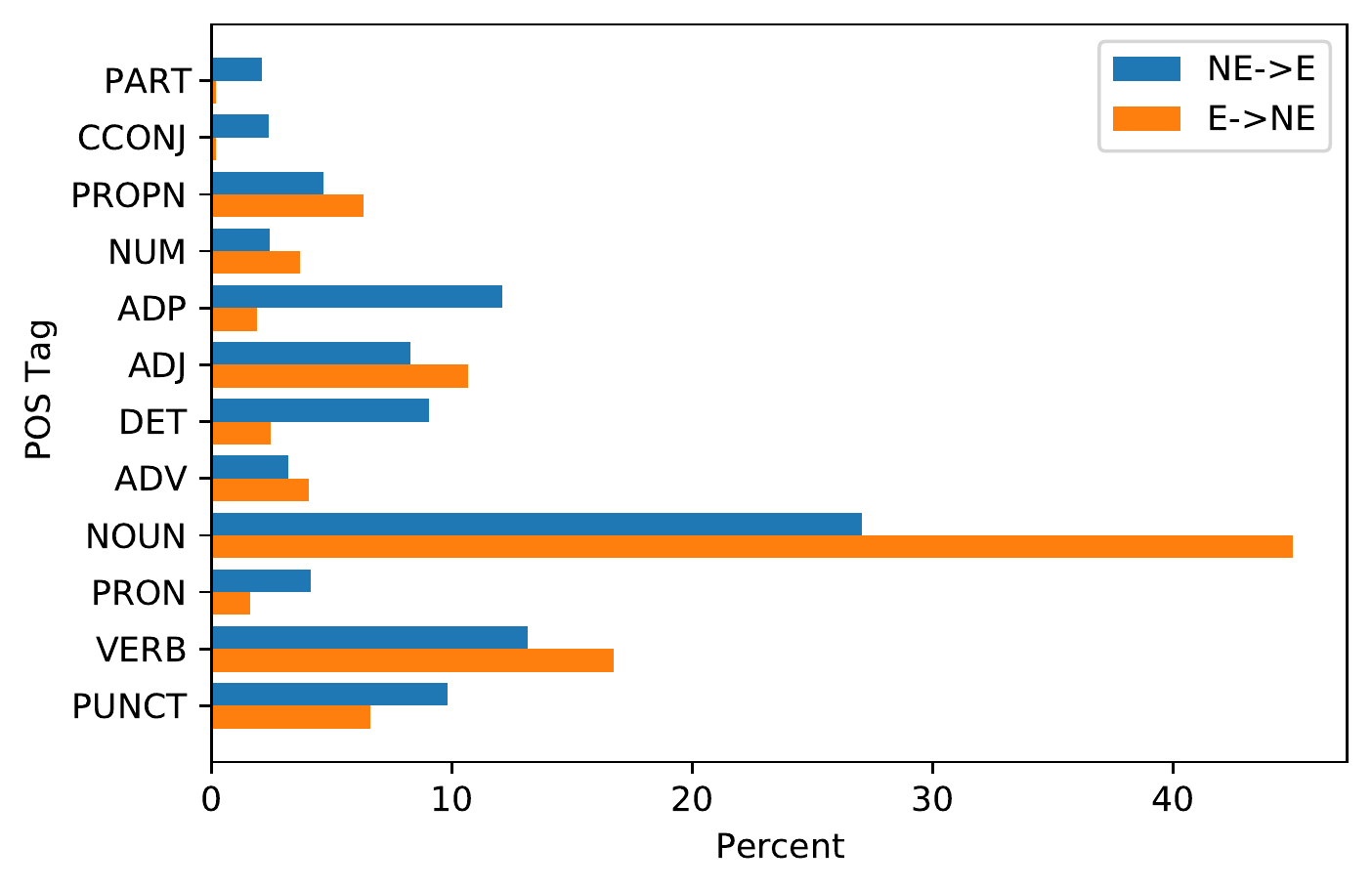} & \includegraphics[scale=0.5]{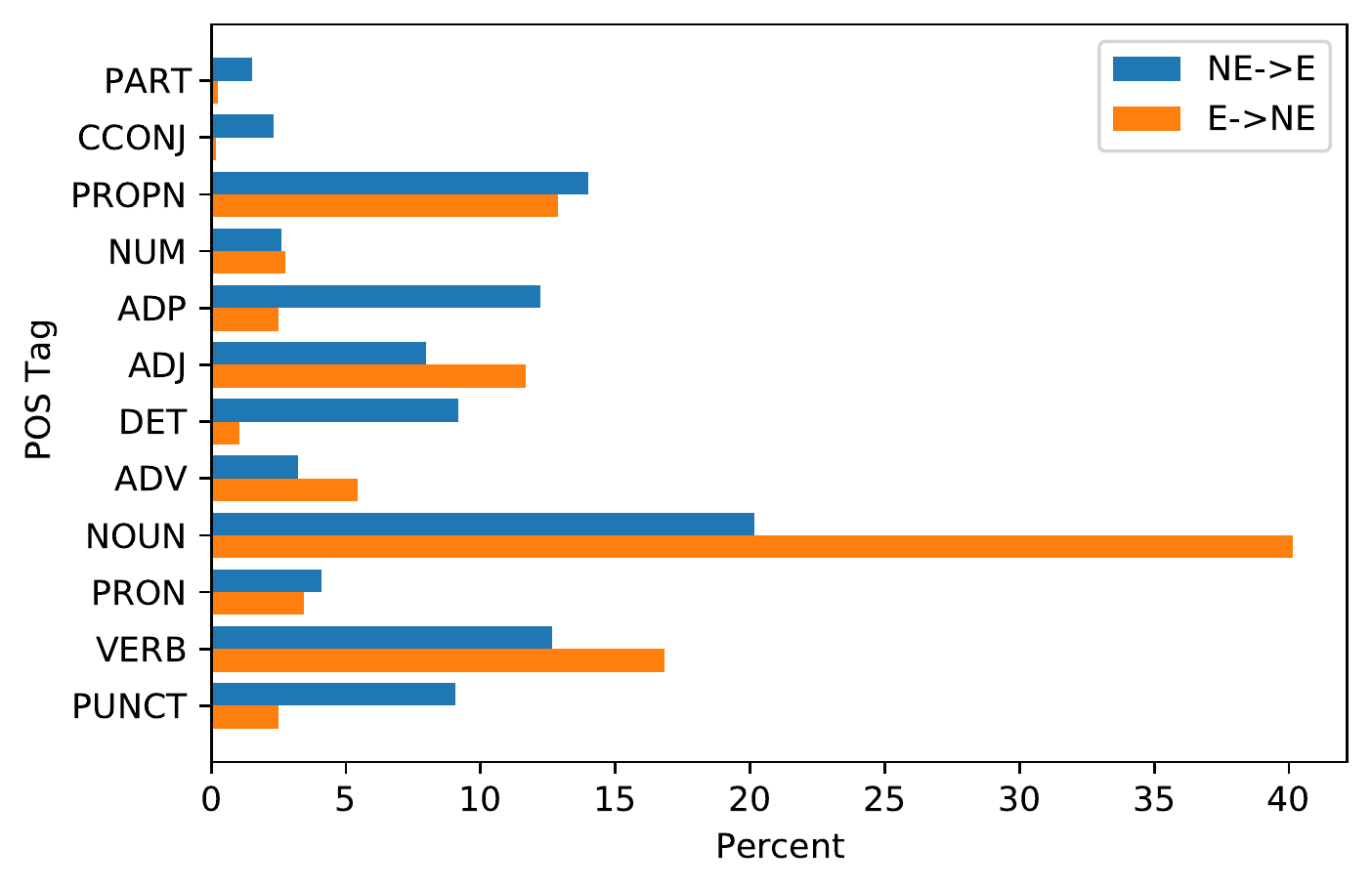}\\
        (c) BERT QNLI & (d) RoBERTa QNLI
    \end{tabular}
    \caption{Barcharts showing how the changes CLOSS makes are distributed among each part of speech tag.}
    \label{figure:poschanges}
\end{figure*}

\subsection{Distribution over Percent of Tokens Changed}
\label{Asect:dist}
Figure \ref{fig:fracchanges} contains histograms showing the distribution of percent of tokens changed over successfully generated counterfactuals. Note that flipping entailment to non-entailment requires fewer changes than the reverse.

\begin{figure*}[]
    \centering
    \begin{tabular}{c c}
        \includegraphics[scale=0.5]{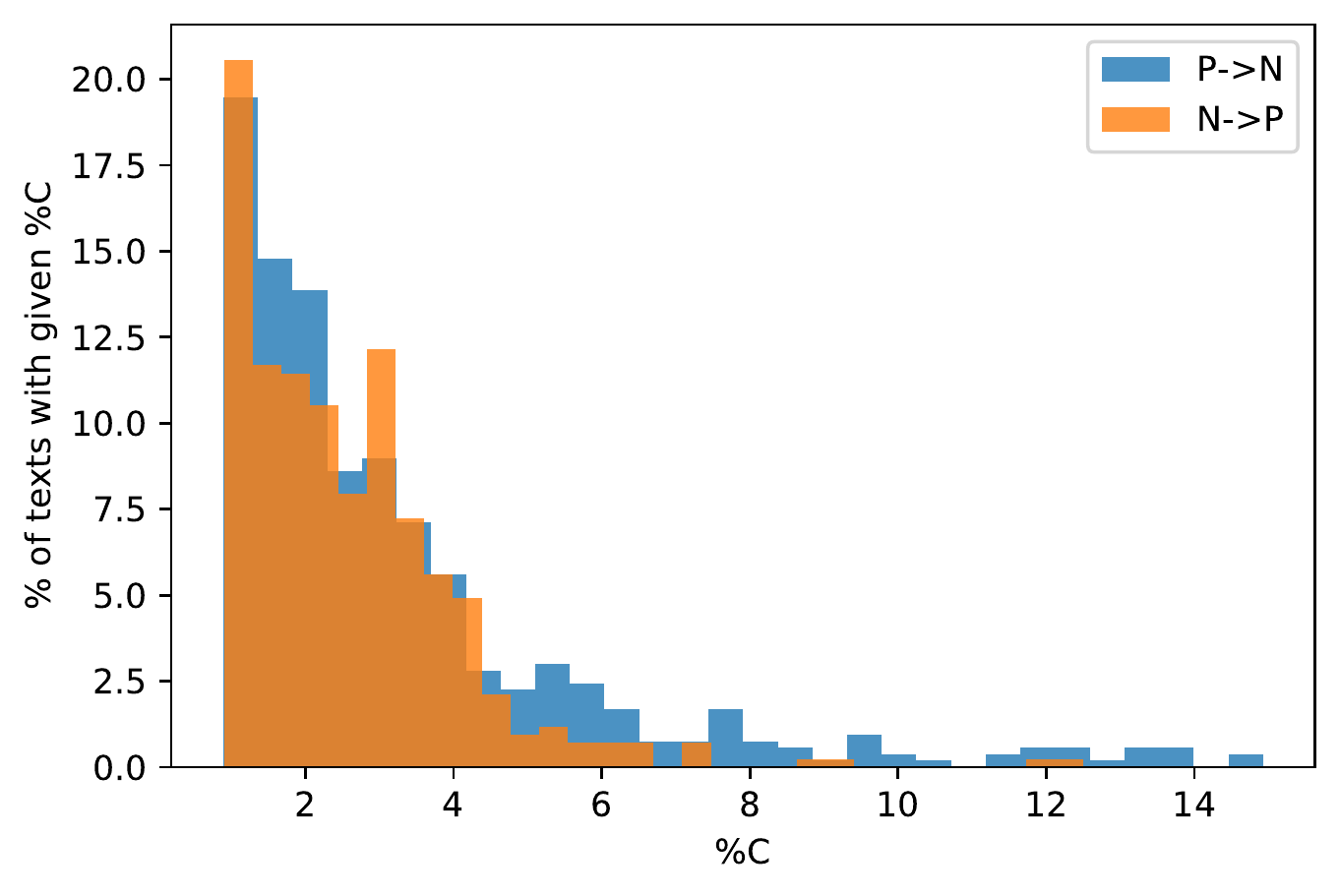} &  \includegraphics[scale=0.5]{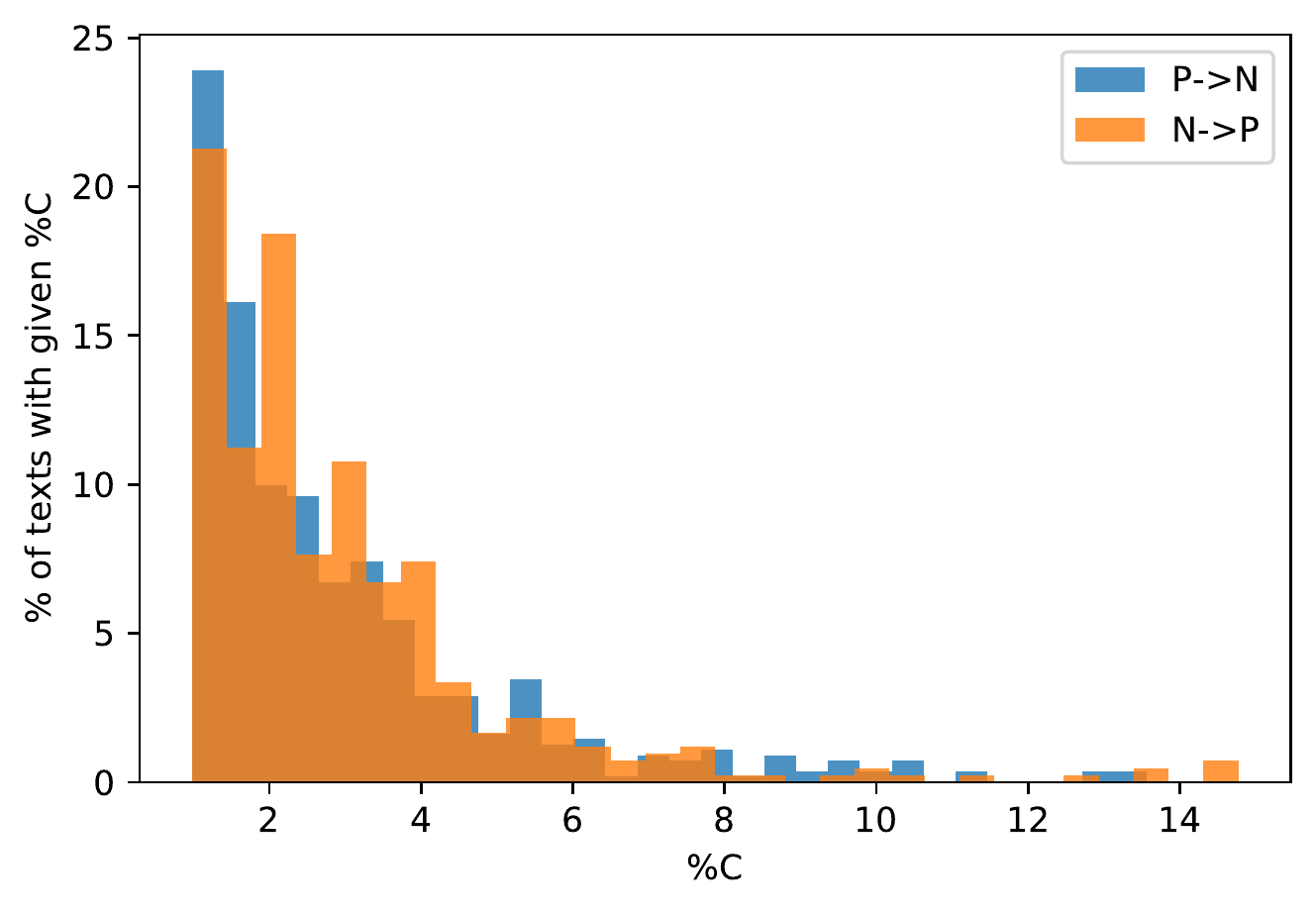}\\
        (a) BERT IMDB & (b) RoBERTa IMDB\\
        \includegraphics[scale=0.5]{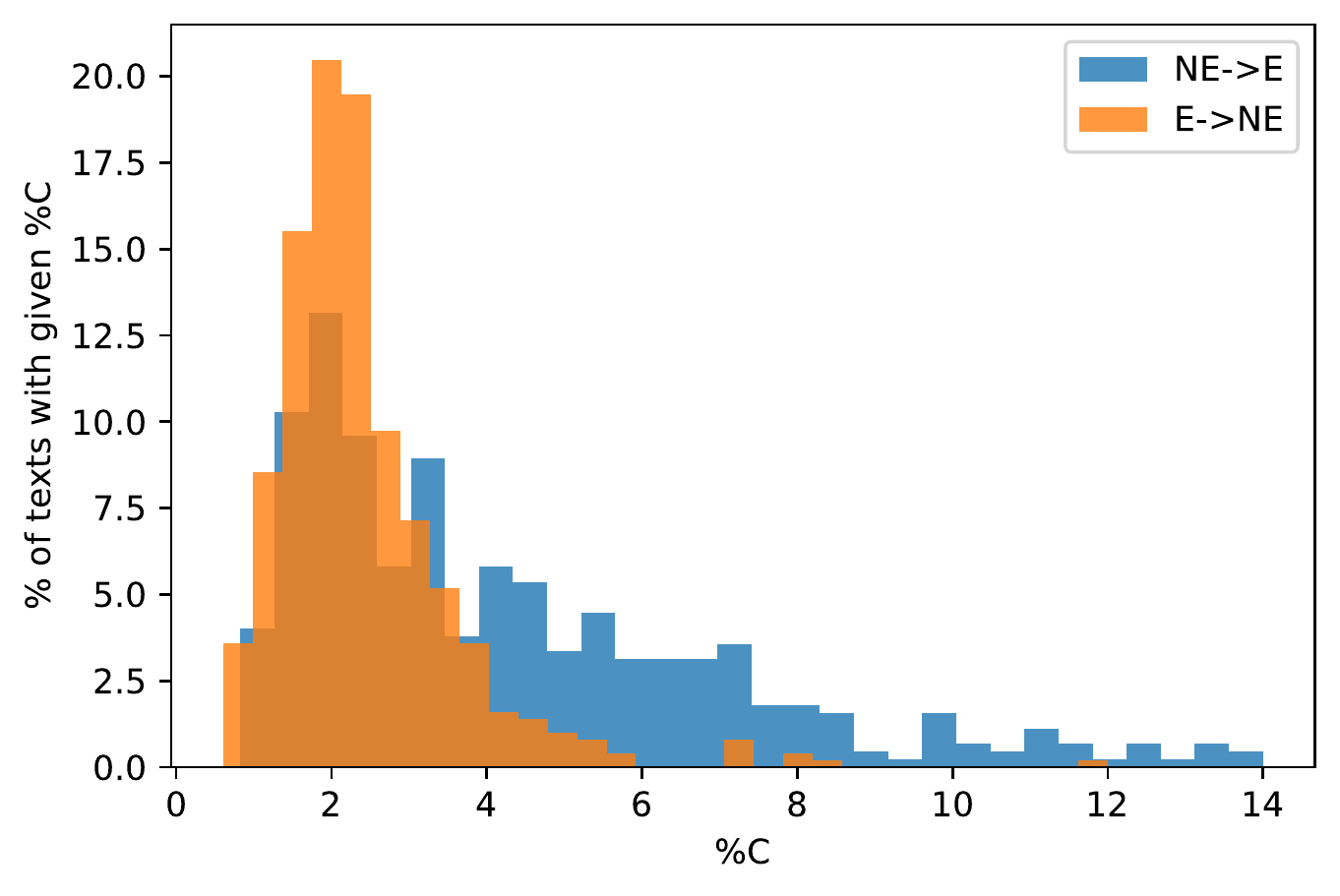} & \includegraphics[scale=0.5]{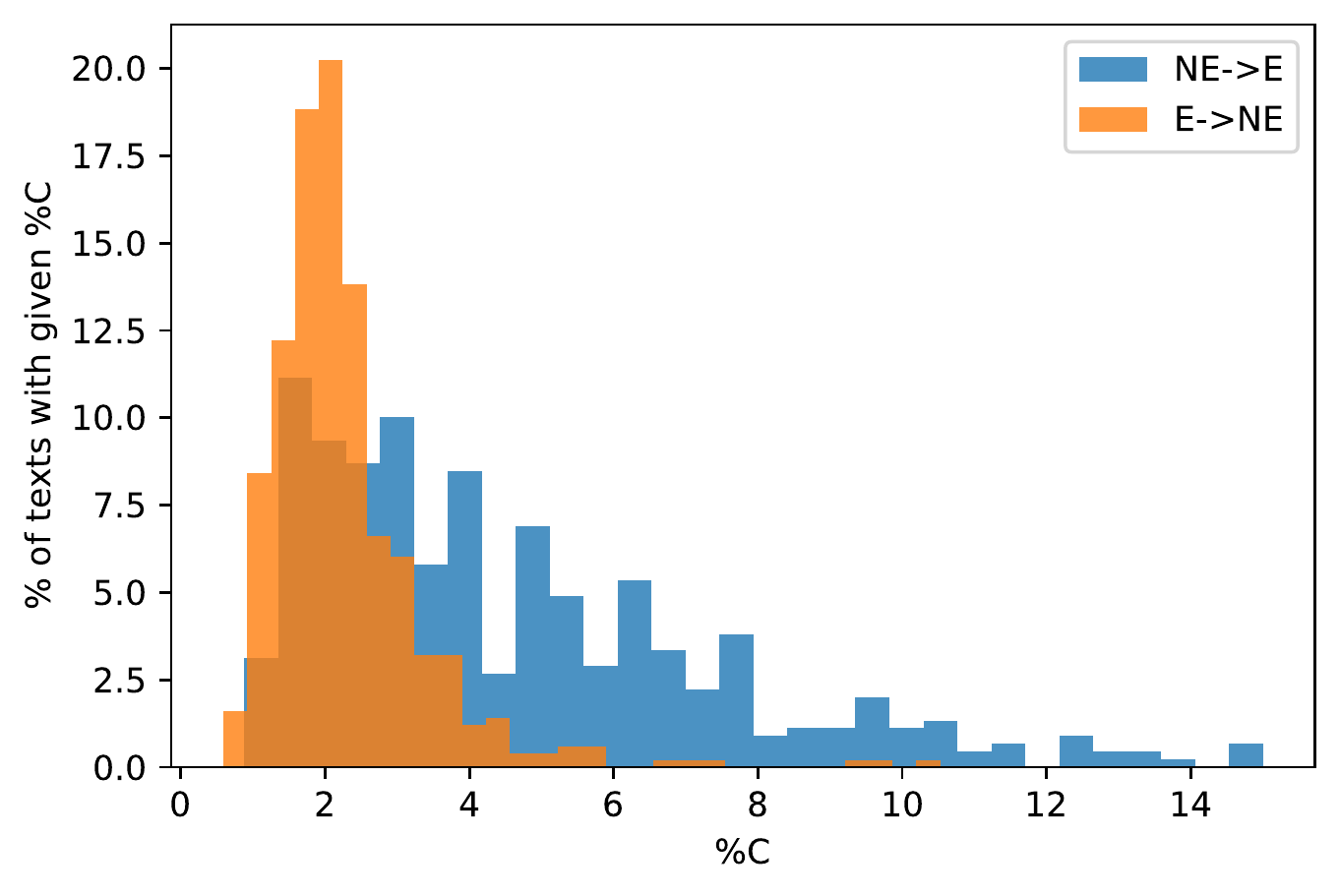}\\
        (c) BERT QNLI & (d) RoBERTa QNLI
    \end{tabular}
    \caption{Histograms showing the distribution of percent of tokens changed over successfully generated counterfactuals.}
    \label{fig:fracchanges}
\end{figure*}

\subsection{Error Analysis}
\label{asect:error}
We explore potential sources of counterfactual generation failure in table \ref{table:erroranalysis} by significantly increasing the computational resources devoted to certain steps of CLOSS and recording the resulting generation failure rate (\%F). 


Even greatly increasing $w$ does not reduce \%F significantly. In comparison, increasing beam width is more effective, especially in regards to IMDB. The most effective interventions are to increase either $K$ or render all tokens salient and scale $w$ in proportion to the associated increase in potential substitutions. Note that when we increase $K$ without changing $w$, the compute spent on estimating Shapley values scales linearly with $K$.

These results suggest failures in the beam search are more of a bottleneck on performance than failing to identify useful substitutions with Shapley values.

However, we can significantly improve performance by increasing both the pool of potential substitutions and the compute spent on estimating Shapley values. This implies many generation failures happen because the pool of potential substitutions the default CLOSS hyperparameters are able to search through does not contain substitutions able to flip the classification.

\begin{table*}
\begin{tabular}{l | cc | cc }
Change & RoBERTa IMDB & BERT IMDB & RoBERTa QNLI & BERT QNLI\\
CLOSS                             & 4.2 & 4.1 & 5.1 & 3.5 \\
Increase beam width               & 0.9 & 0.8 & 2.6 & 3.3 \\
Increase $w$                      & 2.1 & 3.2 & 3.9 & 4.0 \\
Increase $K$                      & \textbf{0.4} & \textbf{0.3} & 1.4 & 2.4 \\
Everything salient, fixed $w$                & 6.2 & 9.3 & 6.4 & 4.0 \\
Everything salient, scale $w$  & 0.53 & 0.93 & \textbf{0.53} & \textbf{0.23}
\end{tabular}

\caption{Shows the impact on failure rate of significantly increasing CLOSS hyperparameters. Beam wdith increases from 15 to 100, w from 5 to 50, K from 30 to 300.}
\label{table:erroranalysis}
\end{table*}

\end{document}